\pdfoutput=1

\documentclass[11pt]{article}

\usepackage[final]{acl}

\usepackage{times}
\usepackage{latexsym}
\usepackage{xspace}

\usepackage[T1]{fontenc}

\usepackage[utf8]{inputenc}

\usepackage{microtype}

\usepackage{inconsolata}

\usepackage{booktabs}

\title{Self-Distillation Bridges Distribution Gap in Language Model Fine-Tuning}

\newcommand{\cadcg}{\textsuperscript{\S}}
\newcommand{\sealab}{\textsuperscript{\textdagger}}
\newcommand{\tencent}{\textsuperscript{\textparagraph}}
\newcommand{\zju}{\textsuperscript{\textdaggerdbl}}

\author{Zhaorui Yang{\cadcg}, Tianyu Pang{\sealab}, Haozhe Feng{\tencent}, Han Wang{\cadcg}\\
\textbf{Wei Chen{\cadcg}, Minfeng Zhu\textsuperscript{\textdaggerdbl\textnormal{\textasteriskcentered}}, Qian Liu\textsuperscript{\textdagger\textnormal{\textasteriskcentered}}}\\ 
{\cadcg}State Key Lab of CAD\&CG, Zhejiang University\\
{\sealab}Sea AI Lab, Singapore \quad{\tencent}Tencent TEG \quad {\zju}Zhejiang University\\
\texttt{minfeng\_zhu@zju.edu.cn, liuqian@sea.com}
}

\usepackage{graphicx} 
\usepackage{xcolor}
\usepackage{multirow}

\newcommand{\increase}[1]{\textcolor[HTML]{006400}{\scriptsize $\uparrow$ #1}}
\newcommand{\decrease}[1]{\textcolor[HTML]{8B0000}{\scriptsize $\downarrow$ #1}}

\usepackage{tikz}    
\usetikzlibrary{shapes.misc}    
  
\newcommand\RoundBox[2][]{    
  \begin{tikzpicture}    
    \node[draw, rectangle, rounded corners, inner sep=4pt, fill=yellow!20,    
          text width=0.9\columnwidth, align=left, #1] (char) {#2};    
  \end{tikzpicture}    
}

\definecolor{examplecolor}{RGB}{253,232,213}

\newcommand\BigExampleBox[2][]{      
    \begin{tikzpicture}  
      \node[draw, rectangle, rounded corners, inner sep=4pt, fill=examplecolor,  
            text width=\dimexpr\textwidth-8pt\relax, align=left, #1] (char) {#2};  
    \end{tikzpicture}  
}

\usepackage{amsmath,amsfonts,bm}

\def\eqref#1{equation~\ref{#1}}

\def\1{\bm{1}}

\DeclareMathAlphabet{\mathsfit}{\encodingdefault}{\sfdefault}{m}{sl}
\SetMathAlphabet{\mathsfit}{bold}{\encodingdefault}{\sfdefault}{bx}{n}

\begin{document}
\maketitle

\let\thefootnote\relax\footnotetext{\textsuperscript{\textasteriskcentered}Corresponding authors}

\begin{abstract}
The surge in Large Language Models (LLMs) has revolutionized natural language processing, but fine-tuning them for specific tasks often encounters challenges in balancing performance and preserving general instruction-following abilities.
In this paper, we posit that the distribution gap between task datasets and the LLMs serves as the primary underlying cause.
To address the problem, we introduce Self-Distillation Fine-Tuning (SDFT), a novel approach that bridges the distribution gap by guiding fine-tuning with a distilled dataset generated by the model itself to match its original distribution.
Experimental results on the Llama-2-chat model across various benchmarks demonstrate that SDFT effectively mitigates catastrophic forgetting while achieving comparable or superior performance on downstream tasks compared to the vanilla fine-tuning.
Moreover, SDFT demonstrates the potential to maintain the helpfulness and safety alignment of LLMs.
Our code is available at \url{https://github.com/sail-sg/sdft}.
\end{abstract}
\section{Introduction}
\label{introduction}
In recent years, the development of Large Language Models (LLMs) has emerged as one of the most groundbreaking advancements in Natural Language Processing (NLP). LLMs such as GPT-3~\citep{gpt3} and PaLM~\citep{palm} have revolutionized the field by leveraging massive textual corpora during pre-training, enabling them to achieve remarkable few-shot performance across a wide range of tasks.
The introduction of Supervised Fine-Tuning (SFT)~\citep{instructgpt,flan} has further propelled the capabilities of LLMs, particularly in enhancing their instruction-following abilities.
\begin{figure}[tb]
    \centering
    \includegraphics[width=\columnwidth]{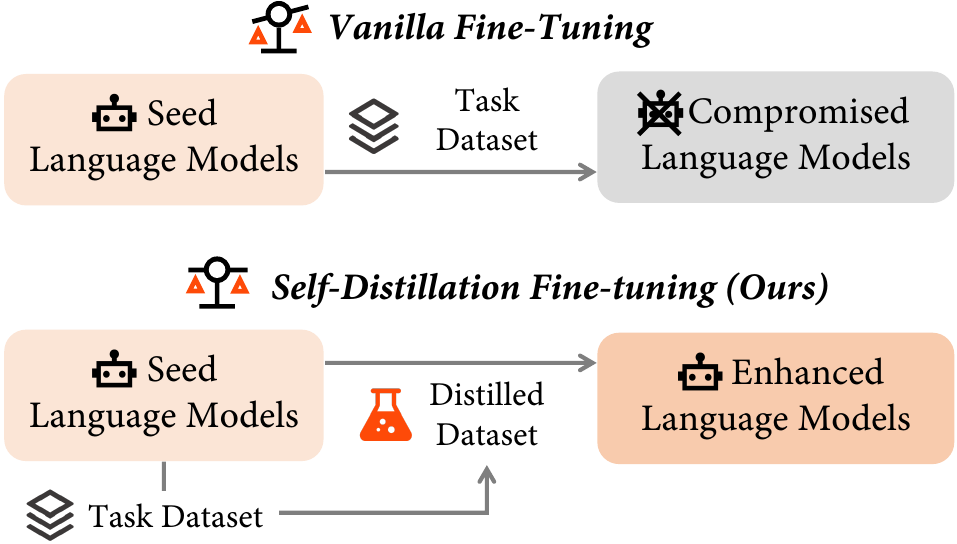}
    \caption{Unlike vanilla fine-tuning, which may compromise seed LMs, our proposed self-distillation fine-tuning (SDFT) approach enhances seed LMs with improved downstream task performance while largely maintaining broad capabilities already learned.}
    \label{fig:intro}
\end{figure}

Interestingly, even when starting with the same base LLM~\citep{llama2,qwen}, minor variations in the supervised dataset can lead to significant differences in model performance~\citep{lima, tulu}.
Consequently, the open-source community has witnessed rapid growth in the diversity of LLM variants, incorporating various SFT datasets and techniques, thereby enhancing their usefulness and accessibility.

However, SFT typically prioritizes improving general instruction-following abilities, suggesting that LLMs with SFT might face challenges in specific downstream tasks. As a result, repurposing these models as Seed Language Models (seed LMs) for subsequent fine-tuning tailored to specific downstream tasks has emerged as an appealing approach.
While the approach seems promising, our preliminary study reveals the challenge of simultaneously enhancing task-specific performance and preserving general instruction-following abilities through vanilla fine-tuning, primarily due to the issue of catastrophic forgetting.
Echoing our findings, recent studies have highlighted that fine-tuning, even with benign datasets, can compromise the safety of seed LMs~\citep{qi2023fine, shallowalign, zhan2023removing, pelrine2023exploiting}.
As evidenced, fine-tuning methods aimed at mitigating catastrophic forgetting are still absent.

\begin{figure*}[tb]
    \centering
    \includegraphics[width=\textwidth]{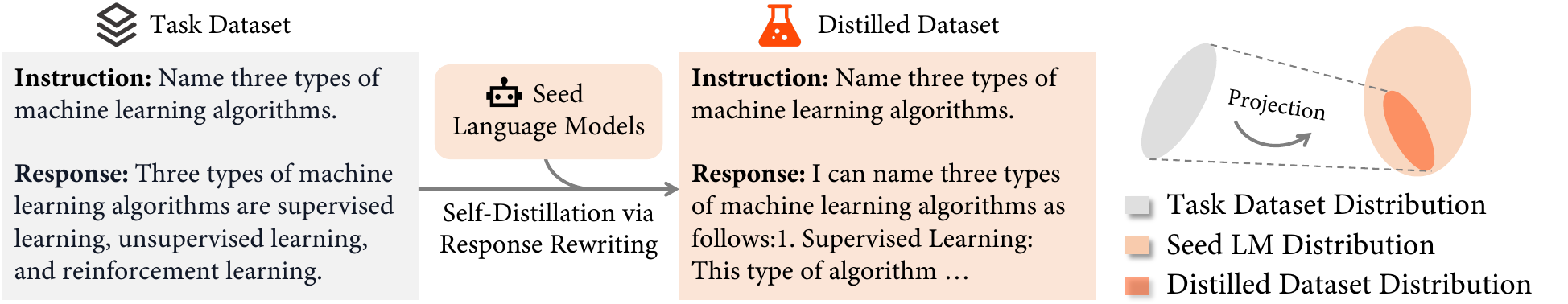}
    \caption{\textbf{Left:} An illustration of a generated distilled response that demonstrates a reduced distribution shift relative to the seed LLM. \textbf{Right:} The diminished distribution shift contributes to a moderate parameter shift, thereby alleviating the issue of catastrophic forgetting.}
    \label{fig:demo}
\end{figure*}

In this paper, we propose a novel fine-tuning method, Self-Distillation Fine-Tuning (SDFT), to mitigate catastrophic forgetting during fine-tuning.
We hypothesize that catastrophic forgetting stems from the distribution gap between the task dataset and the seed LMs.
To address the issue, as shown in Figure~\ref{fig:intro}, SDFT first prompts the seed LM to generate responses that uphold semantic equivalence with the original responses present in the task dataset, resulting in the distilled dataset.
A representative example of rewriting is depicted in Figure~\ref{fig:demo}.
After rewriting, the self-generated responses serve as surrogate targets during subsequent fine-tuning.
Through the approach, SDFT inherently maintains the original distribution, avoiding distribution shift and thereby preserving capabilities.

We systematically evaluate SDFT by comparing its performance against that of vanilla fine-tuning and the seed LM across a variety of benchmarks. These benchmarks encompass: (1) diverse downstream tasks, including \textit{mathematical reasoning}, \textit{tool using} and \textit{code generation}; (2) assessments of general helpfulness and safety alignment. Results on all benchmarks demonstrate the superiority of SDFT compared to vanilla fine-tuning.
For instance, vanilla fine-tuning on the OpenFunctions dataset~\citep{openfunction} leads to a significant decrease in pass@1 on the HumanEval benchmark~\citep{humaneval} from $13.4$ to $9.8$, constituting a decline of $27\%$. In contrast, SDFT not only mitigates this degradation, but also marginally enhances the accuracy to $15.2$. 
The in-depth analysis of our method indicates that increasing the proportion of distilled dataset for fine-tuning leads to a decrease in catastrophic forgetting, thereby confirming that SDFT mitigates catastrophic forgetting by bridging the distribution gap.

\section{Related Work}
\label{relatedworks}

\paragraph{Fine-Tuning}
Fine-tuning is a prevalent strategy for improving the performance of models on downstream tasks, as demonstrated in domains including coding \citep{codellama, wizardcoder}, arithmetic \citep{wizardmath}, healthcare \citep{genegpt} and finance \citep{financellm}. Vanilla fine-tuning directly maximizes the log-likelihood of target responses. Similar to our work, Self-Play Fine-tuning~\citep{self_play_fine_tuning2024} employs the identical LLM as both generator and discriminator, steering the model to prefer annotated response over generated outputs. As the LLM’s distribution ultimately converges with that of the training data, the method does not alleviate forgetting during fine-tuning.

\paragraph{Continual Learning}
Fine-tuning enables models to adapt to new data distributions, improving their efficacy on downstream tasks. However, this process can lead to the loss of previously acquired knowledge, an issue known as catastrophic forgetting \citep{catastrophicforgetting}. A related domain is continual learning \citep{ewc, gem}, which seeks to enable models to acquire new knowledge while mitigating such forgetting. Traditional methods often depend on the preservation of historical data for replay \citep{scialom2022fine, luo2023empirical}, the computation of parameter importance \citep{ewc, mas}, or the assignment of distinct neurons to different tasks \citep{packnet}. However, fine-tuning LLMs is particularly challenging due to their extensive parameter and task space, compounded by the frequent unavailability of original training datasets, which diminishes the feasibility of these established techniques \citep{ewc, gem, scialom2022fine}. Although recent research \citep{luo2023empirical, scialom2022fine} highlights the significance of continual learning for LMs, there are scant feasible solutions for LLMs. In this paper, we conduct a comprehensive evaluation of the catastrophic forgetting issue during the fine-tuning of LLMs and propose a simple yet effective strategy specifically designed for LLMs.

\paragraph{Alignment}
As the capabilities of Large Language Models (LLMs) expand, so does the potential for generating toxic content, engendering significant safety concerns \citep{redteam1, redteam2}. In response, various strategies have been proposed to align LLMs with human ethical standards and prevent the generation of toxic content. Prevalent methods including instruction tuning \citep{DBLP:conf/nips/Ouyang0JAWMZASR22, llama2}, reinforcement learning from human feedback \citep{DBLP:conf/nips/Ouyang0JAWMZASR22, bai2022training}, and self-alignment techniques \citep{selfalign}. Employing these alignment techniques, LLMs strike a dedicate tradeoff between utility and safety \citep{exaggeratedsafety, qi2023fine}. While these methods have demonstrated efficacy in safety alignment, they do not cover further risks that arise from fine-tuning. Recent research reveals that even fine-tuning with benign data can lead to compromised safety \citep{qi2023fine, shallowalign, zhan2023removing, pelrine2023exploiting}. Our proposed strategy can effectively mitigate such safety degradation.  

\paragraph{Prompting Based Learning}
Recently, the use of prompting in LLMs to generate responses for model training has garnered significant interest. Approaches like self-instruct~\cite{selfinstruct} and WizardLM~\cite{wizardlm} utilize the generated responses for supervised fine-tuning, with the latter employing GPT-4 as the generator. Other methods, such as Self-Refine~\cite{selfrefine} and Self-Reward~\cite{selfreward}, use the responses as feedback to iteratively refine the model's outputs. In contrast, our work introduces a novel perspective by leveraging the responses to bridge the distribution gap and address the catastrophic forgetting issue during the fine-tuning process.
\section{Method}
In this section, we begin by outlining the process of fine-tuning, followed by the introduction of our proposed self-distillation fine-tuning method and its implementation details.

\subsection{Fine-tuning LLMs}
\label{sec:formulate}

While LLMs demonstrate remarkable proficiency across various tasks, they often encounter limitations when it comes to downstream tasks that necessitate fine-tuning. Specifically, we refer to a LM in need of further fine-tuning as seed LM, denoted as $f$ and parameterized by $\theta$.
The seed LM typically undergoes general SFT, indicating its capacity to map any natural language instruction $x \in X$ contextualized by the task description $c \in C$, to its corresponding output $y \in Y$.
\begin{equation}
    f_{\theta} : C \times X \rightarrow Y.
\end{equation}
 
The fine-tuning process of the seed LM can be outlined as follows: for the target task $t$ with context $c^t$, each task example $(x^t, y^t)$ is utilized to update the model parameters. 
This update aims at minimizing the disparity between the data distribution and the LM distribution, as expressed below:

\begin{equation}
    L_\text{FT}(\theta)=-\log f_\theta (y^t \mid c^t,x^t),
\end{equation}

which seeks to minimize the negative log likelihood of the target output $y^t$ given the context $c^t$ and input $x^t$, with respect to the model parameters $\theta$.
$L_\text{FT}$ converges when the generated response $\hat{y}$ matches $y^t$, i.e., the distribution of fine-tuned LM aligns with the task dataset distribution.

\subsection{Self-Distillation Fine-Tuning}

As the distribution of the seed LM converges towards that of the task dataset, it naturally enhances performance on target tasks. However, vanilla fine-tuning is susceptible to catastrophic forgetting in general instruction-following capabilities and safety alignment.

To address this issue, we propose \textbf{S}elf-\textbf{D}istillation \textbf{F}ine-\textbf{T}uning (SDFT) to better align the distribution of the task dataset with that of the seed LM.

As depicted in Figure~\ref{fig:demo}, the initial step of SDFT involves prompting the seed LM to rewrite the original response $y^t$ into $\Tilde{y}$:
\begin{equation}
    \Tilde{y} \sim f_{\theta}(y \mid c^t,x^t,y^t).
\end{equation}
This step marks the primary distinction between our method and vanilla fine-tuning, as it involves mapping the original response into a response within the seed LM's distribution. To accomplish the rewriting, we utilize a self-distillation template, which imposes minimal requirements on the seed LM, simply requiring it to adhere to our directive for rewriting responses. The exact specifications of this prompt are elaborated later.

Next, to ensure the quality of the distilled responses, we employ simple heuristics to evaluate the distilled response. For instance, in math reasoning problems, we extract the final answer from the distilled response $\Tilde{y}$ and compare it with the one from the original response $y^t$. Otherwise, we keep the original response. We formalize this conditional selection process as:
\begin{equation}
    \Tilde{y}' =
    \begin{cases}
        \Tilde{y} & \text{if Extract} (\Tilde{y}) = y^t, \\
        y^t & \text{otherwise.}
    \end{cases}
\end{equation}
Finally, the distilled response is used as a replacement for the original response $y^{t}$ for fine-tuning, i.e., the loss becomes:
\begin{equation}
    L_\text{SDFT}(\theta)=-\log f_\theta (\Tilde{y}' \mid c_t,x_t).
\end{equation}
Hence, the distribution gap is mitigated by utilizing the distilled dataset instead of the task dataset, as depicted on the right side of Figure~\ref{fig:demo}.

\subsection{Distillation Template}
\label{sec:dtemplate}

In our work, the distillation template plays a crucial role. Designed to be task-independent, it can be applied seamlessly across various tasks without requiring modification. Within this framework, the template designates the original response within the task dataset as the ``reference answer'' and guides the model to generate a response accordingly. The template employed in the majority of our experiments is illustrated in Figure~\ref{fig:template}. When dealing with datasets involving math reasoning, we slightly adjust the template to better accommodate the reasoning process. Further details about these templates can be found in Appendix~\ref{app:templates}.

\begin{figure}
    \centering
    \RoundBox{
    Below are an instruction that describes a task along with a reference answer. Using the reference answer as a guide, write your own response.
    
    \vspace{\baselineskip}  
    \#\#\# Instruction:
    
    \{instruction\}
    
    \vspace{\baselineskip}
    \#\#\# Reference Answer:
    
    \{original response\}
    
    \vspace{\baselineskip}
    \#\#\# Response:
    }
    \caption{The distillation template used in most of our experiments. It designates the original response as ``reference answer'' and prompts the model to generate a response using the reference answer as a guide.}
    \label{fig:template}
\end{figure}

\section{Experiments}
\label{sec:experiments}

In this section, we begin by presenting the dataset employed for fine-tuning and evaluation purposes. Following that, we conduct a comparative analysis of the experimental results obtained from vanilla fine-tuning and our proposed SDFT approach across various tasks, encompassing mathematical reasoning, code generation, and tool using. Finally, we assess the impact of both methods on safety, general knowledge, and helpfulness.

\subsection{Experimental Setup}
We utilize the Llama-2-7b-chat model~\citep{llama2} as the seed LM in most of our experiments, except where explicitly stated otherwise.
Due to limited computation resources, we utilize the Low Rank Adaptation (LoRA) technique~\citep{lora} during both vanilla fine-tuning and our proposed SDFT.

To ensure fair comparison, we maintain consistency in nearly all hyperparameters for both methods. For datasets comprising more than $10,000$ examples, we randomly select $2,000$ examples for fine-tuning to ensure comparability in size across most datasets. 
For the OpenHermes dataset, we randomly select $20,000$ examples to validate the effect of SDFT with larger, mixed dataset. More experimental details can be found in Appendix~\ref{app:details}.

\subsection{Datasets for Fine-tuning and Evaluation}

We fine-tune the seed LM on a variety of datasets, including those for both single-task and multi-task scenarios.
We then evaluate the performance of both the seed model and the fine-tuned models across diverse tasks.
The datasets for fine-tuning and evaluation are categorized as follows:

\begin{table*}[tb]
\centering
\small
\begin{tabular}{ll*{4}{l}}  
\toprule
\textbf{Method} & \textbf{Dataset} & \textbf{OpenFunctions} & \textbf{GSM8K} & \textbf{HumanEval} & \textbf{Average} \\
\midrule
Seed LM & \textemdash & 19.6 & 29.4 & 13.4  & 20.8 \\
\midrule
\multirow{3}{*}{Vanilla FT} & OpenFunctions & 34.8 &  21.5  & ~~9.8  & 22.0  \\
                          & GSM8K & 17.9  & 31.9  & 12.2  & 20.7  \\
                          & MagiCoder & ~~3.6  & 23.2  & 18.9  & 15.2  \\
\midrule
\multirow{3}{*}{SDFT (Ours)} & OpenFunctions & 36.6 \increase{1.8} &  29.1 \increase{7.6} & 15.2 \increase{5.4} & 27.0 \increase{5.0} \\
                          & GSM8K & 17.9 \increase{0.0} & 34.4 \increase{2.5} & 14.6 \increase{2.4} & 22.3 \increase{1.6} \\
                          & MagiCoder & ~~8.0 \increase{5.4} & 24.9 \increase{1.7} & 18.3 \decrease{0.6} & 17.1 \increase{1.9} \\
\bottomrule
\end{tabular}
\caption{Evaluation results on downstream tasks. The vanilla fine-tuning improves performance on the target task but generally at the expense of tasks that were already performing well. SDFT mitigates the forgetting and can achieve comparable or superior performance on all kinds of tasks.}
\label{tab:performance}
\end{table*}
\newcommand{\cell}[3]{#1 $\rightarrow$ #2  (\increase{#3})} 

\begin{table*}[t]   
\centering
\small  
\begin{tabular}{lcccc}    
\toprule    
\textbf{Dataset for FT} & \textbf{Raw Safe Rate} & \textbf{Jailbreak Safe Rate} & \textbf{AlpacaEval Win Rate} \\
\midrule    
Seed LM & 99.81 & 88.85 & 66.04 \\
\midrule
OpenFunctions & \cell{98.27}{99.23}{~~0.96} & \cell{87.31}{94.42}{~~7.11} & \cell{35.49}{67.66}{32.17} \\
GSM8K & \cell{82.12}{87.12}{~~5.00} & \cell{54.81}{65.58}{10.77} & \cell{23.38}{66.73}{43.35} \\
MagiCoder & \cell{96.73}{97.88}{~~1.15} & \cell{83.65}{88.65}{~~5.00} & 76.52 $\rightarrow$ 76.09 (\decrease{~~0.43}) \\
\bottomrule
\end{tabular}
\caption{Assessment of Safety and General Helpfulness. Results are displayed in the format: \textbf{Vanilla FT $\rightarrow$ SDFT}. Vanilla fine-tuning leads to notable degradation in safety and general helpfulness, while SDFT maintains strong alignment after fine-tuning.}
\label{tab:degradation}
\end{table*}

\begin{table*}[htb]
\centering
\small
\begin{tabular}{ll*{6}{l}}  
\toprule
\textbf{Method} & \textbf{Dataset} & \textbf{Raw Safe Rate} & \textbf{Jailbreak Safe Rate} & \textbf{Win Rate}\\
\midrule
Seed LM & \textemdash & 99.81 & 88.85 & 66.04 \\
\midrule
\multirow{3}{*}{Vanilla FT} & Alpaca & 86.54 & 52.69 & 27.62 \\
                          & Dolly & 81.73 & 26.54 & 22.09 \\
                          & LIMA  & 81.35 & 58.08 & 41.34 \\
                          & OpenHermes & 91.54 & 61.54 & 65.28 \\
                          
\midrule
\multirow{3}{*}{SDFT (Ours)} & Alpaca & 96.15 \increase{9.6} & 86.15 \increase{33.5} & 65.07 \increase{37.5} \\
                          & Dolly & 96.35 \increase{14.6} & 72.69 \increase{46.2} & 61.60 \increase{39.5} \\
                          & LIMA & 94.42 \increase{13.1} & 78.08 \increase{20.0} & 59.38 \increase{18.0} \\
                          & OpenHermes & 95.96 \increase{4.42} & 87.50 \increase{25.96} & 72.91 \increase{7.63} \\
\bottomrule
\end{tabular}
\caption{Evaluation results after fine-tuning on multitask instruction following datasets.}
\label{tab:multitask}
\end{table*}

\begin{figure*}[t]
    \centering
    \includegraphics[width=\textwidth]{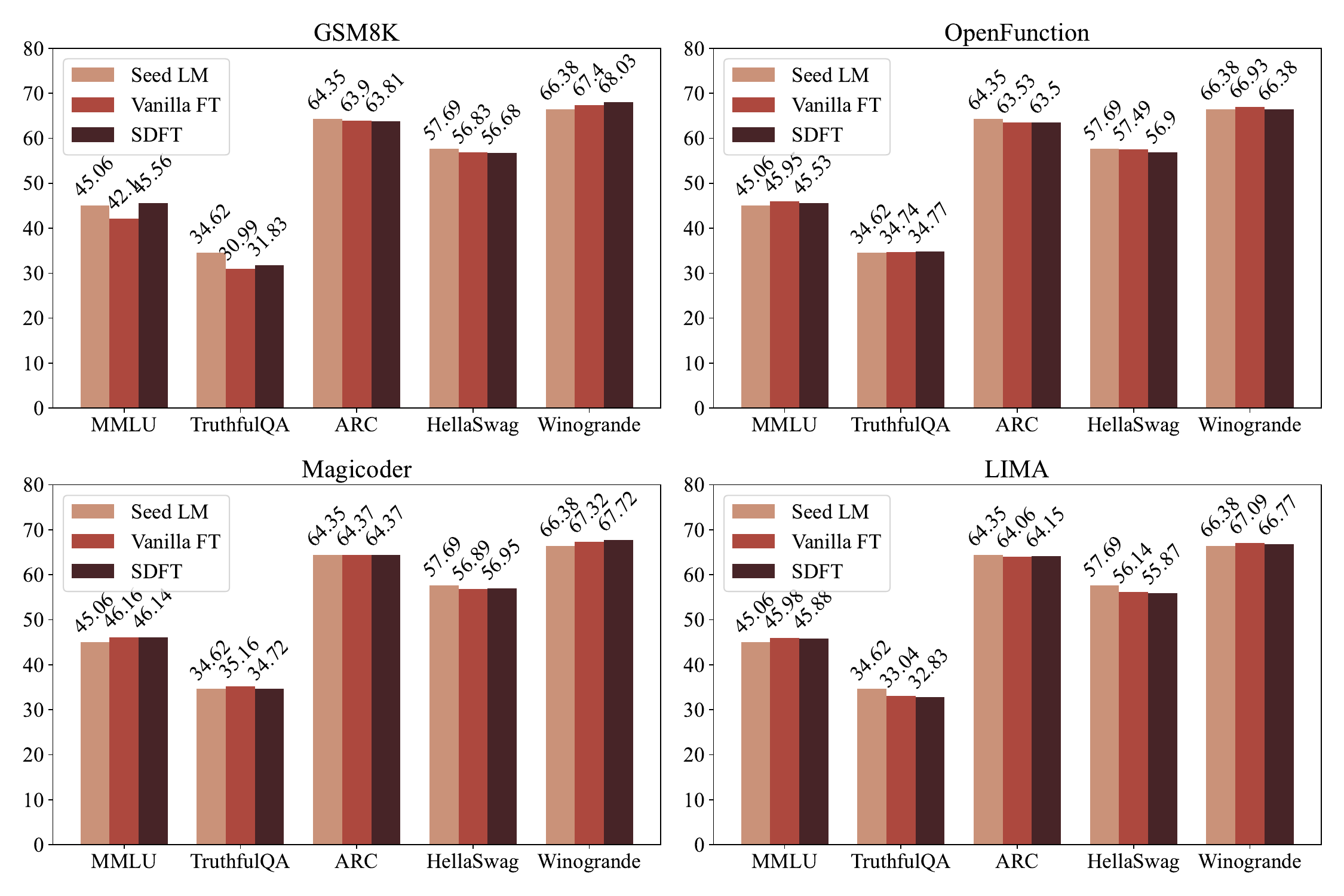}
    \caption{Performance comparisons of models on general knowledge benchmarks after fine-tuning on each dataset, as reported in the OpenLLM Leaderboard. Fine-tuning on these datasets demonstrates a marginal effect on the models' general knowledge.}
    \label{fig:general}
\end{figure*}

\textbf{Single-task datasets.} For single-task datasets, we explore boosting the mathematical reasoning, tool using, and code generation capabilities of LMs during fine-tuning. The mathematical reasoning capabilities are improved using the GSM8K dataset~\citep{gsm8k}, which comprises 8.8k high-quality arithmetic word problems designed at grade school level.
The tool using proficiency is assessed by leveraging function-calling datasets such as the Gorilla OpenFunctions dataset~\citep{openfunction}. Additionally, code generation skills are boosted using the MagiCoder dataset~\citep{magicoder}, while evaluation is conducted using the HumanEval dataset~\citep{humaneval}.

\textbf{Multi-task datasets.}
We use four high-quality datasets to assess the efficacy of our approach within multi-task fine-tuning scenarios: Alpaca~\citep{alpaca}, Dolly~\citep{dolly} and LIMA~\citep{lima}. The Alpaca dataset encompasses a variety of tasks, including arithmetic, coding, and question-answering. It was generated using the Self-Instruct method~\cite{selfinstruct} via the text-davinci-003 model. 
The Dolly dataset is composed of seven distinct tasks, such as open question \& answer, information extraction, and summarization.
The LIMA dataset covers a broad range of topics and was curated from multiple sources.
The OpenHermes dataset consists of primarily GPT-4 generated data from a variety of public datasets, with filtering to remove refusals.

\textbf{Safety evaluation.} We utilize the harmful behavior instructions from the Advbench dataset~\cite{advbench} for evaluation, assessing the safety of models' outputs through keyword matching following \citet{qi2023fine}. We define the proportion of safe responses as \textbf{Raw Safe Rate}. Additionally, we simulate jailbreaking attempts by appending adversarial suffixes to instructions as illustrated in \citet{advbench}. The safe rate under this condition is referred to as \textbf{Jailbreak Safe Rate}.
 
\textbf{Helpfulness evaluation.} We employ AlpacaEval~\cite{alpaca_eval} to evaluate the helpfulness of various models. This tool includes a dataset and associated evaluation metrics that facilitate the comparison of generated outputs with the responses from Text-Davinci-003, across a diverse set of 805 detailed instructions sourced from multiple datasets. We report the win rate, which is the proportion of instances where the responses are favored over those produced by Text-Davinci-003, as judged by GPT-4.

\textbf{Knowledge evaluation.} LMs' general knowledge was assessed through evaluations using benchmarks from the OpenLLM Leaderboard, specifically MMLU~\cite{mmlu}, TruthfulQA~\cite{truthfulqa}, ARC~\cite{arc}, HellaSwag~\cite{hellaswag}, and Winogrande~\cite{winogrande}. These datasets provide a measure of the models' factual and commonsense knowledge spanning a variety of domains.

\subsection{SDFT Achieves Better Results on Downstream Tasks}
Table~\ref{tab:performance} presents the results of fine-tuning on three downstream tasks. The results indicate that while vanilla fine-tuning can enhance the model's efficacy on target tasks, it also leads to a significant decline in performance on other tasks. For example, as depicted in the table’s first row, fine-tuning with the OpenFunctions dataset results in a diminished coding capability of the model, decreasing from $13.41$ to $9.76$. A similar decline is observed in mathematical reasoning abilities, where accuracy on the GSM8K dataset drops from $29.42$ to $21.53$. 

Furthermore, the proposed SDFT can effectively mitigate this performance degradation. In the cited instance, the model retains its mathematical reasoning proficiency, achieving an accuracy of $29.11$, closely aligned with the seed model’s performance ($29.42$). For coding performance evaluated on HumanEval, there is a marginal improvement, with the performance rising to $15.24$ from the seed model's $13.41$.When focusing on the target task, SDFT also outperforms vanilla fine-tuning, delivering an accuracy of $36.61$ compared to $34.82$.

\subsection{SDFT Preserves Alignment}
Fine-tuning on the majority of datasets has been demonstrated to lead to a significant decrease in both safety alignment and general helpfulness, as highlighted by the findings in Table~\ref{tab:degradation}. For instance, following fine-tuning on the GSM8K dataset, the safe rate decreases from $99.81$ to $82.12$, the jailbreak safe rate drops from $88.85$ to $54.81$, and the win rate on AlpacaEval diminishes from $66.04$ to $23.38$. In contrast, our proposed SDFT approach effectively mitigates this decline, improving the raw safe rate and jailbreak safe rate by $5$ and $11$, respectively. Notably, there is a slight increase in the win rate compared to the seed model, with a score of $66.73$ versus $66.03$.

Table~\ref{tab:multitask} presents evaluation results after fine-tuning on instruction following datasets that contain multiple tasks. As the target tasks of these datasets are unspecified, we focus our evaluation on safety and general helpfulness after fine-tuning. In line with the patterns noted in Table~\ref{tab:degradation}, fine-tuning on Alpaca, Dolly and LIMA typically leads to a marked reduction in both safety and helpfulness metrics. We observe a pronounced decline in all three metrics, with each declining by roughly $20$. In contrast, our proposed SDFT method effectively mitigates this reduction, limiting the decrease to under $10$. Similarly, vanilla fine-tuning on the OpenHermes~\cite{openhermes} dataset results in diminished safety alignment. In contrast, SDFT effectively mitigates this degradation, enhancing the jailbreak safe rate from $61.54$ to $87.50$.

\subsection{General Knowledge Remains Intact}
Figure~\ref{fig:general} presents results on general knowledge. Although vanilla fine-tuning compromises downstream performance and alignment, models' capabilities in general knowledge are relatively unaffected. For instance, after fine-tuning on the OpenFunctions dataset, the disparity in performance between fine-tuned model and seed LM is less than $1$. This is also observed after fine-tuning with SDFT.

\section{Analysis}
\label{sec:analysis}

\begin{figure*}[ht]  
    \centering
    \begin{minipage}{0.48\textwidth}  
        \centering
        \includegraphics[width=\linewidth]{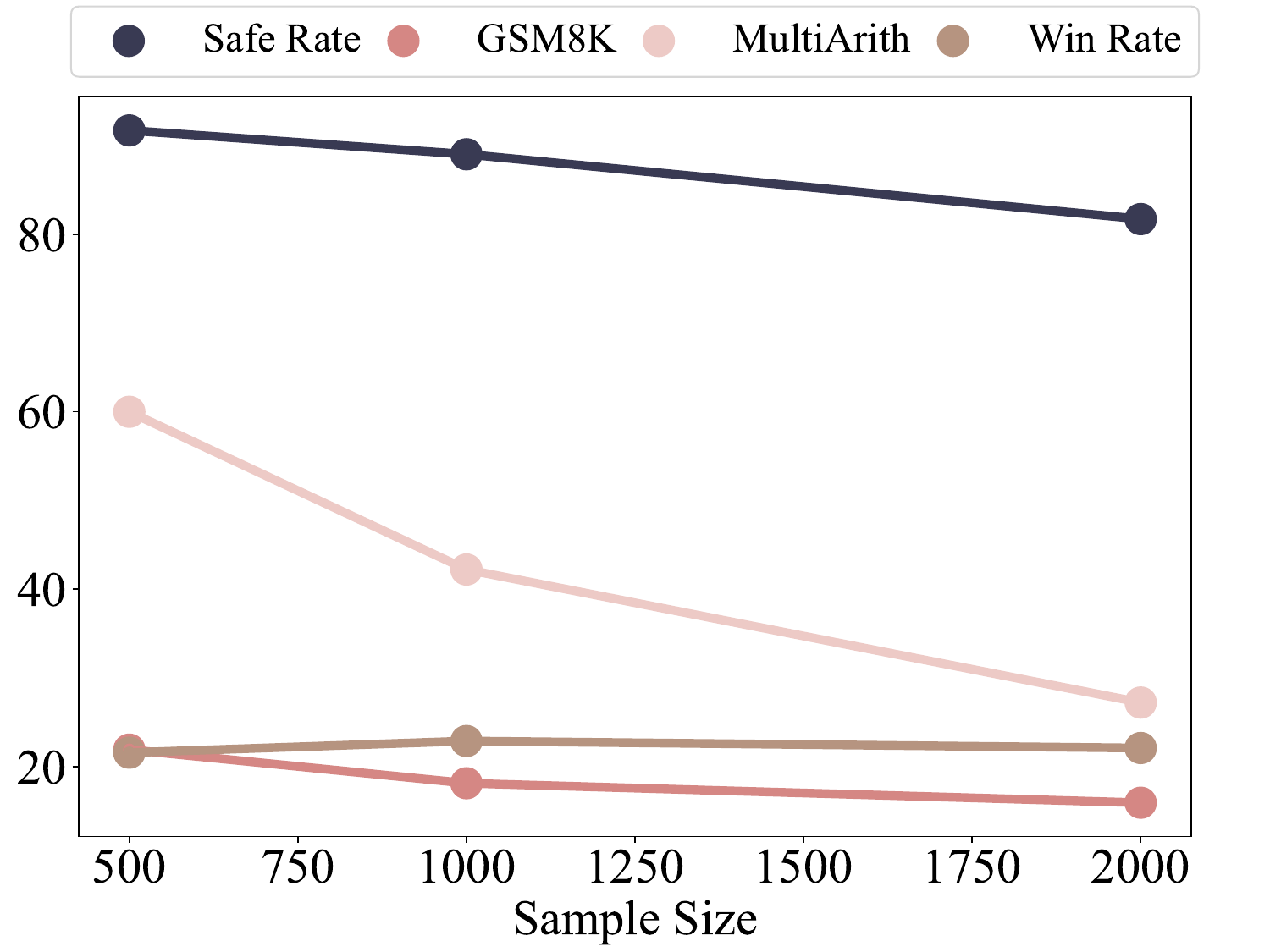}
        \caption{With increasing data for fine-tuning, there\\is a decrease in models' performance across various\\benchmarks, including math, safety alignment and\\instruction following capability.}
        \label{fig:metric_size}  
    \end{minipage} 
    \begin{minipage}{0.48\textwidth}  
        \centering  
        \includegraphics[width=\linewidth]{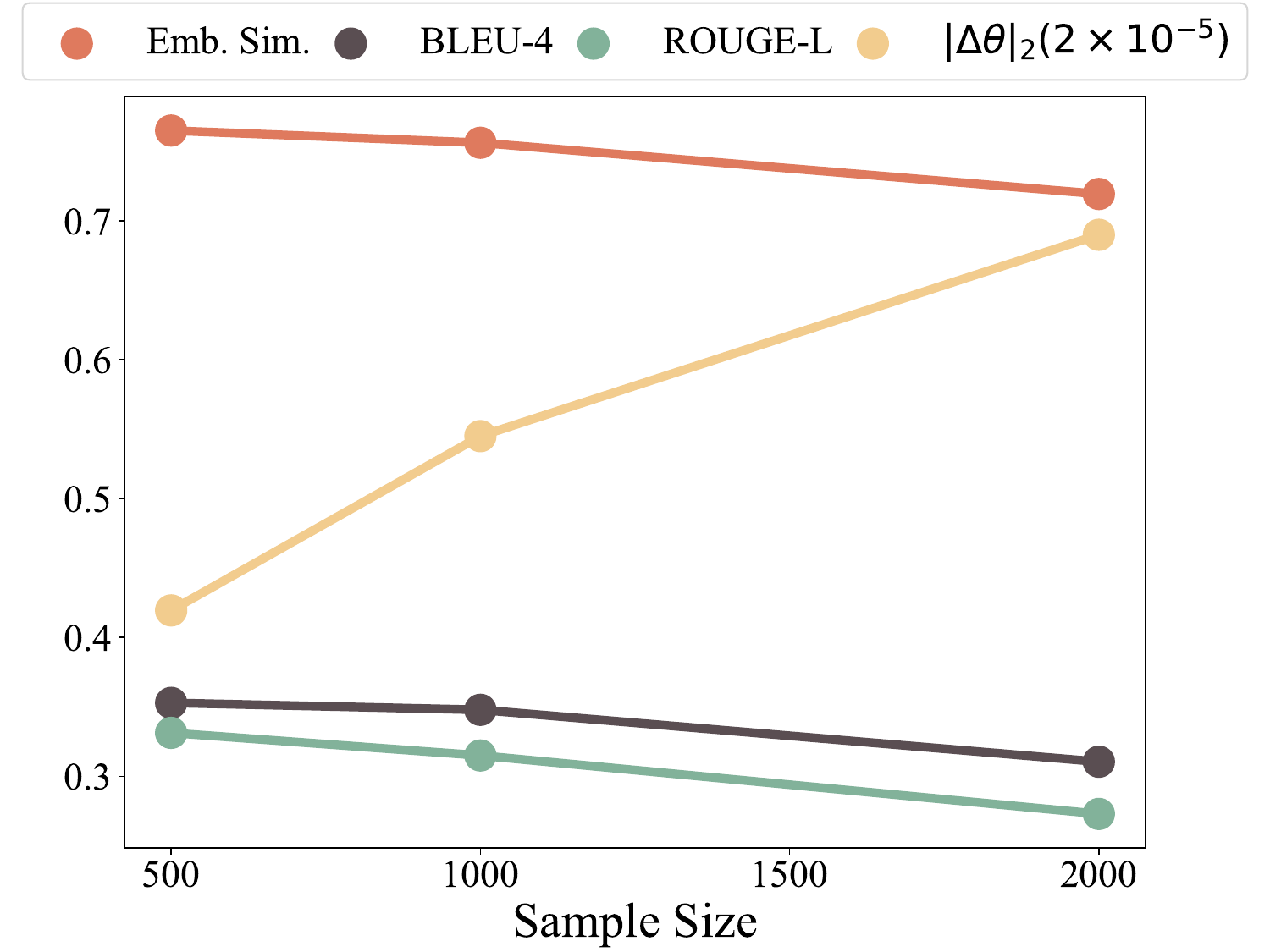}
        \caption{As the sample size increases, BLEU-4, ROUGE-L and embedding similarity all decrease, while parameter shift scale increases, indicating an intensified extent of distribution shift.}
        \label{fig:sim_size}
    \end{minipage}  
\end{figure*}

\begin{figure*}[ht]
    \centering  
    \begin{minipage}{0.48\textwidth}  
        \centering
        \includegraphics[width=\linewidth]{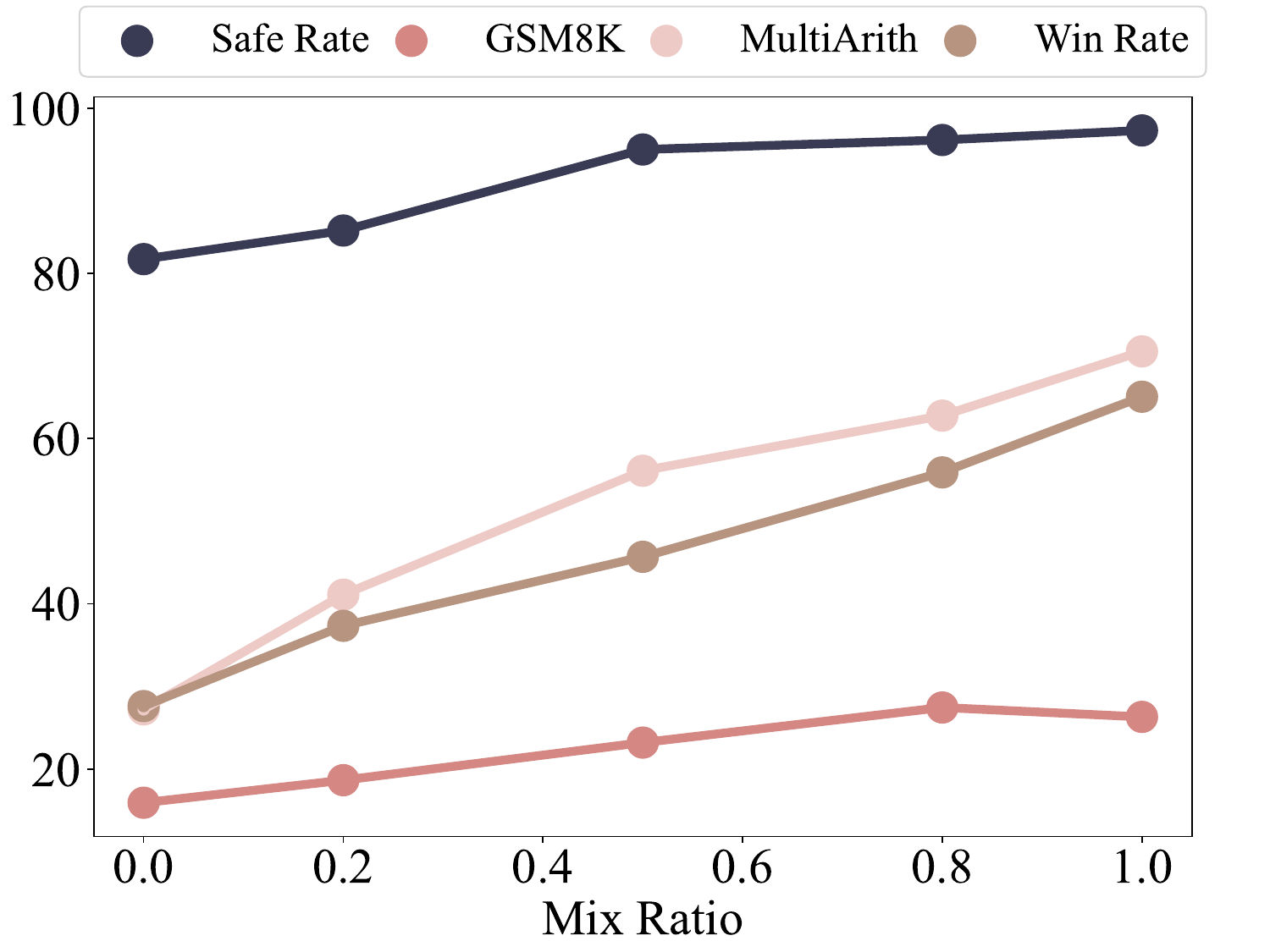}
        \caption{With an increasing mix ratio, there is an\\enhancement in the models' performance across\\various benchmarks.}
        \label{fig:metric_mix}  
    \end{minipage} 
    \begin{minipage}{0.48\textwidth}  
        \centering  
        \includegraphics[width=\linewidth]{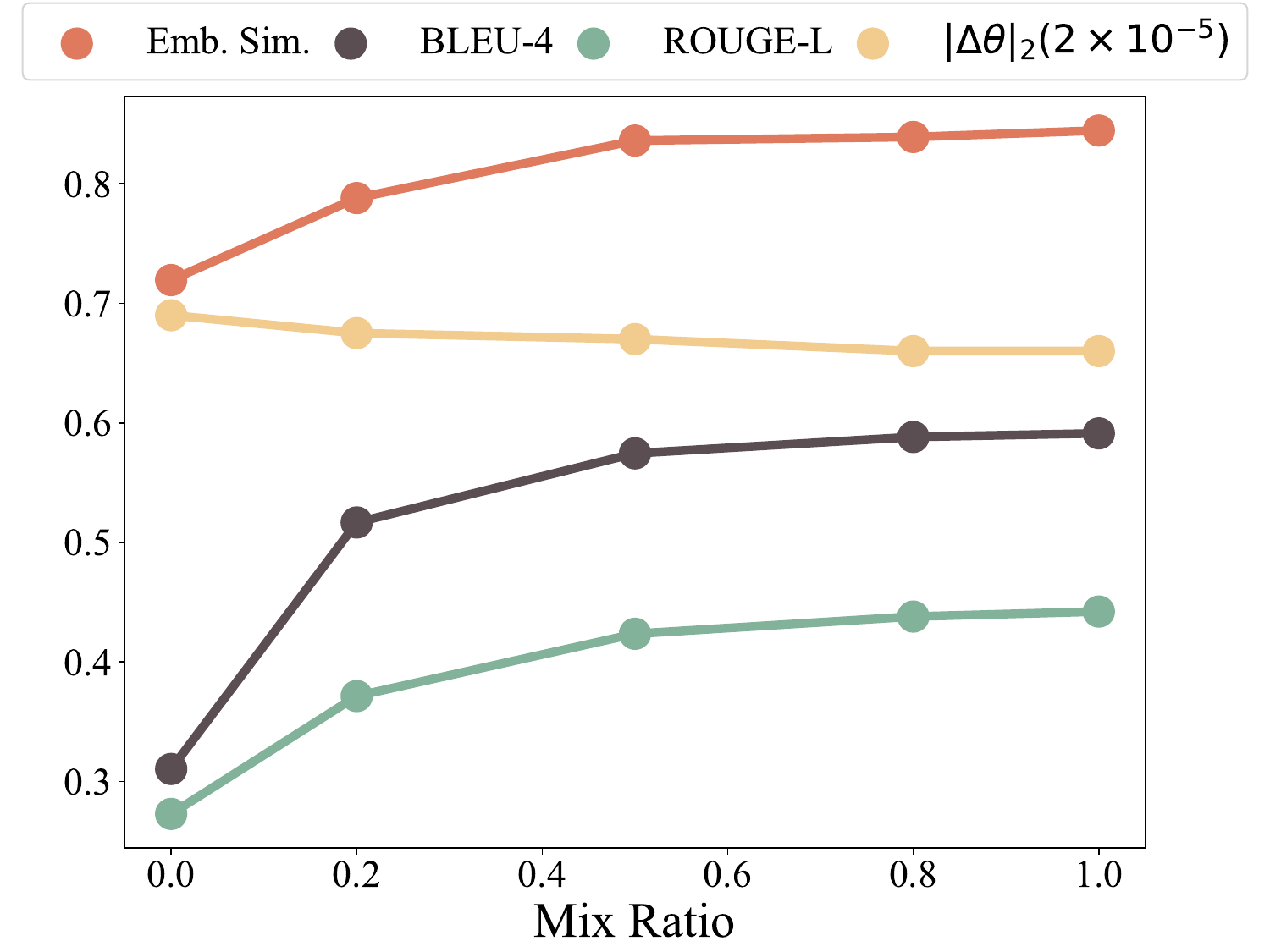}
        \caption{As the mix ratio increases, BLEU-4, ROUGE-L and embedding similarity increase, while parameter shift decreases, indicating reduced distribution shift.}
        \label{fig:sim_mix}
    \end{minipage}  
\end{figure*}  

In this section, we conduct a detailed analysis to understand the impact of distribution shift on catastrophic forgetting. In addition to the evaluation metrics outlined in Section~\ref{sec:experiments}, we incorporate four supplementary metrics to assess the degree of distribution shift. We utilize both the seed model and fine-tuned models to generate responses on the Advbench~\cite{advbench} dataset and engage in a comparative analysis of these responses. 

In particular, we calculate the BLEU-4 and ROUGE-L scores for the fine-tuned models, using the outputs from the seed model as references to evaluate the extent of distribution shift. We also utilize Sentence-BERT~\cite{sentencebert} to derive sentence embeddings and use the cosine similarity between these embeddings following \citet{autocot}. Lastly, we quantify the extent of parameter shift by comparing the updated parameters with those of the seed model, considering their distance as a measure of the parameter shift magnitude. The lower the BLEU-4, ROUGE-L, and embedding similarity scores, the greater the distribution shift. Conversely, the parameter shift is directly proportional to the norm of the parameter changes.

\subsection{Distribution Shift Correlates with Catastrophic Forgetting}
We induce varying degrees of distribution shift too investigate its impact through two approaches: (1) By sampling a diverse quantity of examples for fine-tuning, where an increased number of data points for fine-tuning corresponds to a greater distribution shift. (2) By mixing vanilla fine-tuning with SDFT, which involves substituting distilled samples with original ones. We define mix ratio to represent the proportion of distilled samples employed. A mix ratio of $1$ signifies exclusive use of our SDFT and $0$ denotes vanilla fine-tuning.

Figures~\ref{fig:metric_size} and \ref{fig:sim_size} illustrate the results with varying sample sizes. As the sample size grows, we observe a notable decrease in the BLEU-4, ROUGE-L, and embedding similarity scores, along with an elevation in parameter shift magnitude. This trend implies a heightened degree of distribution shift. Consequently, there is an observable decline in model performance on benchmarks such as GSM8K, MultiArith, Advbench, and AlpacaEval, suggesting intensified catastrophic forgetting.

In a similar vein, Figures \ref{fig:metric_mix} and \ref{fig:sim_mix} present results corresponding to an ascending mix ratio. As this ratio increases, there is an upward trend in the BLEU-4, ROUGE-L, and embedding similarity scores, whereas the scale of parameter shift diminishes, denoting a mitigation in distribution shift. Accordingly, benchmark performance exhibits improvement across the board, signaling a reduction in the severity of catastrophic forgetting.

\begin{figure}[ht]
    \centering
    \includegraphics[width=\columnwidth]{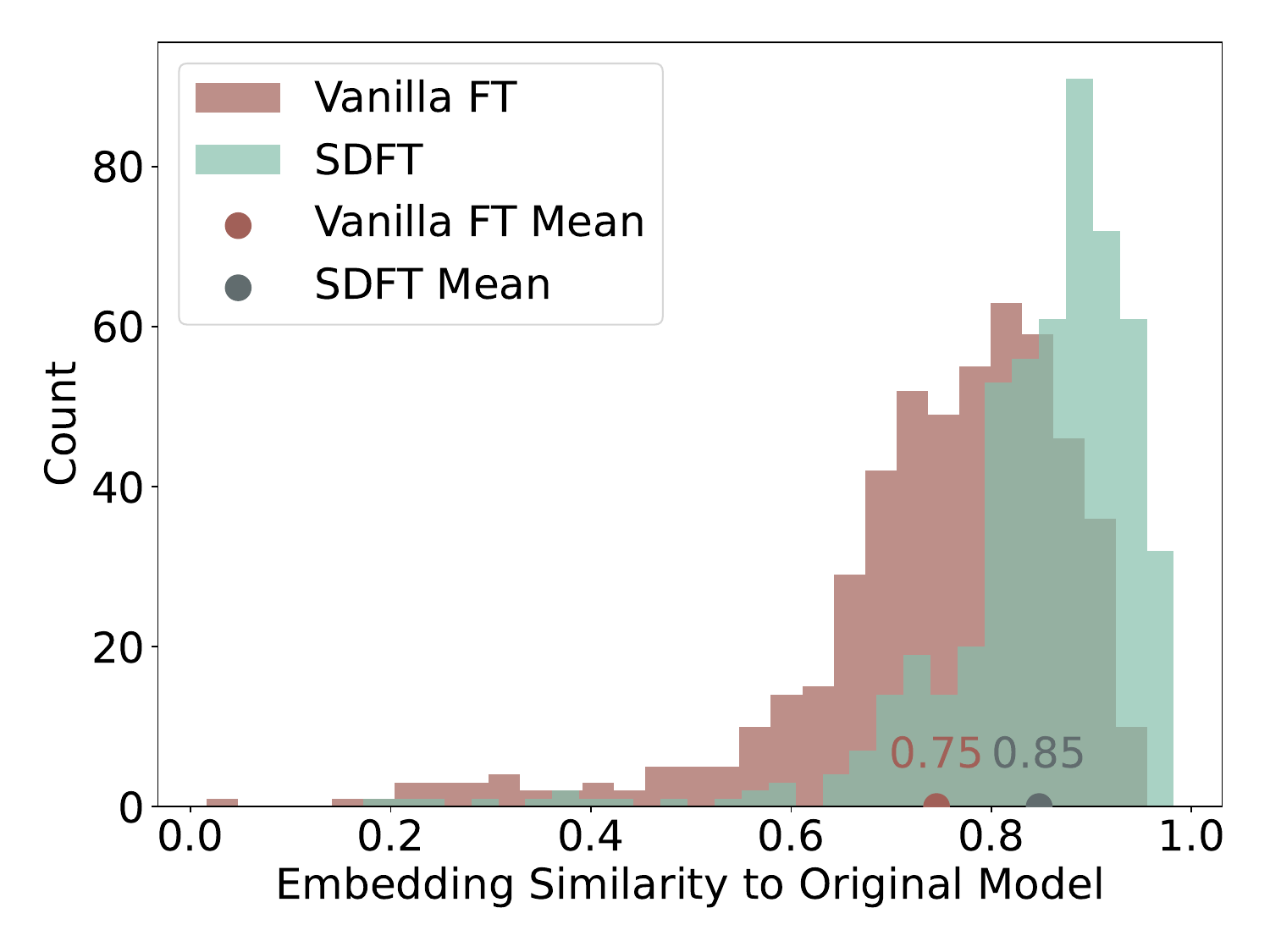}
    \caption{The distribution of embedding similarities after fine-tuning. SDFT results in higher similarity to the original model, indicating reduced distribution shift.}
    \label{fig:similarity}
\end{figure}

Figure~\ref{fig:similarity} illustrates the similarity distribution obtained through both vanilla fine-tuning and our SDFT. Notably, with SDFT model has higher similarity between the fine-tuned model and the seed model, signifying reduced distribution shift.

\subsection{Robustness among Distillation Templates}
We have constructed two templates to investigate the robustness of SDFT. The template illustrated in Figure~\ref{fig:template} is labeled ``Using'', where the phrase ``Using the reference answer as a guide'' is replaced by ``Refer to the reference answer'', with the latter template being termed ``Refer''. Results after fine-tuning with both templates are detailed in Table~\ref{tab:ablation_template}. The performance across diverse benchmarks remains consistent across the templates, demonstrating the robustness of SDFT.

\begin{table*}[ht]  
\centering
\small
\begin{tabular}{ll*{6}{c}}  
\toprule  
\textbf{Dataset for FT} & \textbf{Template} & \textbf{OpenFunctions}  & \textbf{HumanEval} & \textbf{GSM8K} & \textbf{Raw Safe} & \textbf{Jailbreak Safe} & \textbf{Win Rate} \\
\midrule
\multirow{3}{*}{OpenFunctions} & Vanilla FT & 34.82 & ~~9.76 & 21.53 & 98.27 & 87.31 & 35.49 \\
                              & Refer & 35.71 & 13.41 & 27.37 & 98.85 & 89.81 & \textbf{68.45} \\
                              & Using & \textbf{36.61} & \textbf{15.24} & \textbf{29.11} & \textbf{99.23} & \textbf{94.42} & 67.66 \\
\midrule

\multirow{3}{*}{Dolly} & Vanilla FT & ~~8.04 & \textbf{17.07} & 15.92 & 81.73 & 26.54 & 22.09 \\
                       & Refer & \textbf{17.86} & 14.02 & 24.26 & 96.35 & 69.62 & \textbf{61.60} \\
                       & Using & 16.07 & 14.63 & \textbf{26.31} & \textbf{97.31} & \textbf{72.69} & 57.52 \\

\bottomrule
\end{tabular}  
\caption{Ablation studies on distillation template. The performance of SDFT is consistently better than Vanilla FT with different distillation templates.}
\label{tab:ablation_template}
\end{table*}
\begin{table*}[t]
    \centering\small
    \begin{tabular}{l*{6}{l}}
    \toprule
    \textbf{Method} & \textbf{GSM8K} & \textbf{OpenFunctions} & \textbf{HumanEval} & \textbf{Raw Safe} & \textbf{Jailbreak Safe} & \textbf{Win Rate} \\
    \midrule
    \multicolumn{7}{c}{\textit{Dataset for FT: GSM8K}} \\
    Seed LM (7B) & 29.40 & 19.60 & 13.41 & 99.81 & 88.85 & 66.04 \\ 
    Vanilla FT (full) & 34.87 & 5.36 & 13.41 & 84.62 & 37.31 & 23.04 \\
    SDFT (Ours, full) & 35.03 \increase{0.16} & 16.07 \increase{10.71} & 15.85 \increase{2.44} & 88.46 \increase{3.84} & 63.46 \increase{26.15} & 61.19 \increase{38.15} \\
    \midrule
    \multicolumn{7}{c}{\textit{Dataset for FT: GSM8K}} \\
    Seed LM (13B) & 38.06 & 36.61 & 19.51 & 99.81 & 98.85 & 86.75 \\
    Vanilla FT (LoRA) & 44.12 & 19.64 & 17.68 & 94.42 & 88.27 & 40.27 \\
    SDFT (Ours, LoRA) & 45.59 \increase{1.47} & 24.11 \increase{4.47} & 18.28 \increase{0.61} & 97.31 \increase{2.89} & 94.42 \increase{6.15} & 75.93 \increase{35.66} \\
    \midrule
    \multicolumn{7}{c}{\textit{Dataset for FT: OpenFunctions}} \\
    Llama3-8B-Instruct & 81.58 & 41.07 & 59.76 & 95.58 & 94.81 & 75.34 \\ 
    Vanilla FT (LoRA) & 77.79 & 42.86 & 54.27 & 88.85 & 79.81 & 79.75\\
    SDFT (Ours, LoRA) & 79.45 \increase{1.66} & 43.75 \increase{0.89} & 56.10 \increase{1.83} & 92.12 \increase{3.27} & 96.15 \increase{16.34} & 82.24 \increase{2.49}\\
    \bottomrule
    \end{tabular}
    \caption{Evaluation of our SDFT under full fine-tuning on Llama-2-7b-chat, LoRA fine-tuning on Llama-2-13b-chat model, and LoRA fine-tuning on Llama-3-8B-Instruct model using different fine-tuning datasets.}
    \label{tab:full-13b-llama3}
\end{table*}

\subsection{Efficacy of SDFT Across Model Scales and Architectures}
The SDFT approach is not constrained by any specific fine-tuning technique (such as LoRA) or model architecture, enabling its application across both comprehensive fine-tuning processes and other model architectures. To substantiate this claim, we conducted supplementary experiments that included full fine-tuning on Llama-2-7b-chat and LoRA fine-tuning on Llama-2-7b-chat. Additionally, we explored the fine-tuning of the recently unveiled SOTA model, Llama3~\cite{llama3} on the OpenFunctions dataset. The results in Table~\ref{tab:full-13b-llama3} reveal that in all scenarios, SDFT not only consistently outperforms vanilla fine-tuning in the target task but also reduces forgetting across all other tasks, demonstrating its effectiveness.  
\section{Conclusions and Limitations}
In this paper, we perform a systematic evaluation of catastrophic forgetting during the fine-tuning of language models for downstream tasks. Our findings indicate that the distribution shift during fine-tuning can lead to performance degradation in general task capabilities, as well as models' safety alignment and helpfulness. To enhance performance on target task while maintaining LMs' broad capabilities, we propose a plug-and-play strategy, SDFT, to reduce distribution shift and mitigate catastrophic forgetting. Extensive experiments show that SDFT effectively diminishes forgetting and delivers comparable or superior performance to vanilla fine-tuning on targeted tasks. 

Our study is subject to certain limitations. Owing to constraints in computational resources, most of our experiments are based on the Llama-2-7b-chat model with LoRA. Further investigations involving larger models and full fine-tuning remain to be explored. Furthermore, our safety evaluations are limited to the Advbench dataset and fixed adversarial suffixes, leaving the robustness against other jailbreaking strategies for future work.

\section*{Ethics Statement}
Our proposed method SDFT effectively mitigates the issue of catastrophic forgetting during the fine-tuning of language models, including the degradation of safety alignment. Therefore, this process does not entail additional risks.

We utilize a variety of open-source English datasets for training, including Alpaca, Dolly, LIMA, OpenHermes, GSM8K, OpenFunctions, and MagiCoder. The Llama-2-chat model serves as our seed model for training. We acknowledge that there may be inherent biases present within these datasets and the model.    

\section*{Acknowledgements}
This paper is supported by the National Science Foundation of China (62132017, 62302435), Zhejiang Provincial Natural Science Foundation of China (LD24F020011, LQ24F020006) and ``Pioneer and Leading Goose'' R\&D Program of Zhejiang (2024C01167).



\bibliography{custom}

\begin{thebibliography}{53}
\expandafter\ifx\csname natexlab\endcsname\relax\def\natexlab#1{#1}\fi

\bibitem[{Aljundi et~al.(2018)Aljundi, Babiloni, Elhoseiny, Rohrbach, and Tuytelaars}]{mas}
Rahaf Aljundi, Francesca Babiloni, Mohamed Elhoseiny, Marcus Rohrbach, and Tinne Tuytelaars. 2018.
\newblock Memory aware synapses: Learning what (not) to forget.
\newblock In \emph{{European Conference on Computer Vision}}, pages 139--154.

\bibitem[{Bai et~al.(2023)Bai, Bai, Chu, Cui, Dang, Deng, Fan, Ge, Han, Huang, Hui, Ji, Li, Lin, Lin, Liu, Liu, Lu, Lu, Ma, Men, Ren, Ren, Tan, Tan, Tu, Wang, Wang, Wang, Wu, Xu, Xu, Yang, Yang, Yang, Yang, Yao, Yu, Yuan, Yuan, Zhang, Zhang, Zhang, Zhang, Zhou, Zhou, Zhou, and Zhu}]{qwen}
Jinze Bai, Shuai Bai, Yunfei Chu, Zeyu Cui, Kai Dang, Xiaodong Deng, Yang Fan, Wenbin Ge, Yu~Han, Fei Huang, Binyuan Hui, Luo Ji, Mei Li, Junyang Lin, Runji Lin, Dayiheng Liu, Gao Liu, Chengqiang Lu, Keming Lu, Jianxin Ma, Rui Men, Xingzhang Ren, Xuancheng Ren, Chuanqi Tan, Sinan Tan, Jianhong Tu, Peng Wang, Shijie Wang, Wei Wang, Shengguang Wu, Benfeng Xu, Jin Xu, An~Yang, Hao Yang, Jian Yang, Shusheng Yang, Yang Yao, Bowen Yu, Hongyi Yuan, Zheng Yuan, Jianwei Zhang, Xingxuan Zhang, Yichang Zhang, Zhenru Zhang, Chang Zhou, Jingren Zhou, Xiaohuan Zhou, and Tianhang Zhu. 2023.
\newblock Qwen technical report.
\newblock \emph{arXiv preprint arXiv:2309.16609}.

\bibitem[{Bai et~al.(2022)Bai, Jones, Ndousse, Askell, Chen, DasSarma, Drain, Fort, Ganguli, Henighan et~al.}]{bai2022training}
Yuntao Bai, Andy Jones, Kamal Ndousse, Amanda Askell, Anna Chen, Nova DasSarma, Dawn Drain, Stanislav Fort, Deep Ganguli, Tom Henighan, et~al. 2022.
\newblock Training a helpful and harmless assistant with reinforcement learning from human feedback.
\newblock \emph{arXiv preprint arXiv:2204.05862}.

\bibitem[{Bianchi et~al.(2023)Bianchi, Suzgun, Attanasio, R{\"o}ttger, Jurafsky, Hashimoto, and Zou}]{exaggeratedsafety}
Federico Bianchi, Mirac Suzgun, Giuseppe Attanasio, Paul R{\"o}ttger, Dan Jurafsky, Tatsunori Hashimoto, and James Zou. 2023.
\newblock Safety-tuned llamas: Lessons from improving the safety of large language models that follow instructions.
\newblock \emph{arXiv preprint arXiv:2309.07875}.

\bibitem[{Brown et~al.(2020)Brown, Mann, Ryder, Subbiah, Kaplan, Dhariwal, Neelakantan, Shyam, Sastry, Askell et~al.}]{gpt3}
Tom Brown, Benjamin Mann, Nick Ryder, Melanie Subbiah, Jared~D Kaplan, Prafulla Dhariwal, Arvind Neelakantan, Pranav Shyam, Girish Sastry, Amanda Askell, et~al. 2020.
\newblock Language models are few-shot learners.
\newblock In \emph{{Advances in Neural Information Processing Systems}}, volume~33, pages 1877--1901.

\bibitem[{Chen et~al.(2021)Chen, Tworek, Jun, Yuan, Pinto, Kaplan, Edwards, Burda, Joseph, Brockman et~al.}]{humaneval}
Mark Chen, Jerry Tworek, Heewoo Jun, Qiming Yuan, Henrique Ponde de~Oliveira Pinto, Jared Kaplan, Harri Edwards, Yuri Burda, Nicholas Joseph, Greg Brockman, et~al. 2021.
\newblock Evaluating large language models trained on code.
\newblock \emph{arXiv preprint arXiv:2107.03374}.

\bibitem[{Chen et~al.(2024)Chen, Deng, Yuan, Ji, and Gu}]{self_play_fine_tuning2024}
Zixiang Chen, Yihe Deng, Huizhuo Yuan, Kaixuan Ji, and Quanquan Gu. 2024.
\newblock \href {https://doi.org/10.48550/ARXIV.2401.01335} {Self-play fine-tuning converts weak language models to strong language models}.
\newblock \emph{CoRR}, abs/2401.01335.

\bibitem[{Chowdhery et~al.(2023)Chowdhery, Narang, Devlin, Bosma, Mishra, Roberts, Barham, Chung, Sutton, Gehrmann, Schuh, Shi, Tsvyashchenko, Maynez, Rao, Barnes, Tay, Shazeer, Prabhakaran, Reif, Du, Hutchinson, Pope, Bradbury, Austin, Isard, Gur{-}Ari, Yin, Duke, Levskaya, Ghemawat, Dev, Michalewski, Garcia, Misra, Robinson, Fedus, Zhou, Ippolito, Luan, Lim, Zoph, Spiridonov, Sepassi, Dohan, Agrawal, Omernick, Dai, Pillai, Pellat, Lewkowycz, Moreira, Child, Polozov, Lee, Zhou, Wang, Saeta, Diaz, Firat, Catasta, Wei, Meier{-}Hellstern, Eck, Dean, Petrov, and Fiedel}]{palm}
Aakanksha Chowdhery, Sharan Narang, Jacob Devlin, Maarten Bosma, Gaurav Mishra, Adam Roberts, Paul Barham, Hyung~Won Chung, Charles Sutton, Sebastian Gehrmann, Parker Schuh, Kensen Shi, Sasha Tsvyashchenko, Joshua Maynez, Abhishek Rao, Parker Barnes, Yi~Tay, Noam Shazeer, Vinodkumar Prabhakaran, Emily Reif, Nan Du, Ben Hutchinson, Reiner Pope, James Bradbury, Jacob Austin, Michael Isard, Guy Gur{-}Ari, Pengcheng Yin, Toju Duke, Anselm Levskaya, Sanjay Ghemawat, Sunipa Dev, Henryk Michalewski, Xavier Garcia, Vedant Misra, Kevin Robinson, Liam Fedus, Denny Zhou, Daphne Ippolito, David Luan, Hyeontaek Lim, Barret Zoph, Alexander Spiridonov, Ryan Sepassi, David Dohan, Shivani Agrawal, Mark Omernick, Andrew~M. Dai, Thanumalayan~Sankaranarayana Pillai, Marie Pellat, Aitor Lewkowycz, Erica Moreira, Rewon Child, Oleksandr Polozov, Katherine Lee, Zongwei Zhou, Xuezhi Wang, Brennan Saeta, Mark Diaz, Orhan Firat, Michele Catasta, Jason Wei, Kathy Meier{-}Hellstern, Douglas Eck, Jeff Dean, Slav Petrov, and Noah Fiedel.
  2023.
\newblock \href {http://jmlr.org/papers/v24/22-1144.html} {Palm: Scaling language modeling with pathways}.
\newblock \emph{J. Mach. Learn. Res.}, 24:240:1--240:113.

\bibitem[{Chung et~al.(2022)Chung, Hou, Longpre, Zoph, Tay, Fedus, Li, Wang, Dehghani, Brahma et~al.}]{flan}
Hyung~Won Chung, Le~Hou, Shayne Longpre, Barret Zoph, Yi~Tay, William Fedus, Yunxuan Li, Xuezhi Wang, Mostafa Dehghani, Siddhartha Brahma, et~al. 2022.
\newblock Scaling instruction-finetuned language models.
\newblock \emph{arXiv preprint arXiv:2210.11416}.

\bibitem[{Clark et~al.(2018)Clark, Cowhey, Etzioni, Khot, Sabharwal, Schoenick, and Tafjord}]{arc}
Peter Clark, Isaac Cowhey, Oren Etzioni, Tushar Khot, Ashish Sabharwal, Carissa Schoenick, and Oyvind Tafjord. 2018.
\newblock Think you have solved question answering? try arc, the ai2 reasoning challenge.
\newblock \emph{arXiv:1803.05457v1}.

\bibitem[{Cobbe et~al.(2021)Cobbe, Kosaraju, Bavarian, Chen, Jun, Kaiser, Plappert, Tworek, Hilton, Nakano, Hesse, and Schulman}]{gsm8k}
Karl Cobbe, Vineet Kosaraju, Mohammad Bavarian, Mark Chen, Heewoo Jun, Lukasz Kaiser, Matthias Plappert, Jerry Tworek, Jacob Hilton, Reiichiro Nakano, Christopher Hesse, and John Schulman. 2021.
\newblock Training verifiers to solve math word problems.
\newblock \emph{arXiv preprint arXiv:2110.14168}.

\bibitem[{Conover et~al.(2023)Conover, Hayes, Mathur, Xie, Wan, Shah, Ghodsi, Wendell, Zaharia, and Xin}]{dolly}
Mike Conover, Matt Hayes, Ankit Mathur, Jianwei Xie, Jun Wan, Sam Shah, Ali Ghodsi, Patrick Wendell, Matei Zaharia, and Reynold Xin. 2023.
\newblock \href {https://www.databricks.com/blog/2023/04/12/dolly-first-open-commercially-viable-instruction-tuned-llm} {Free dolly: Introducing the world's first truly open instruction-tuned llm}.

\bibitem[{French(1999)}]{catastrophicforgetting}
Robert~M French. 1999.
\newblock Catastrophic forgetting in connectionist networks.
\newblock \emph{Trends in cognitive sciences}, 3(4):128--135.

\bibitem[{Ganguli et~al.(2022)Ganguli, Lovitt, Kernion, Askell, Bai, Kadavath, Mann, Perez, Schiefer, Ndousse et~al.}]{redteam2}
Deep Ganguli, Liane Lovitt, Jackson Kernion, Amanda Askell, Yuntao Bai, Saurav Kadavath, Ben Mann, Ethan Perez, Nicholas Schiefer, Kamal Ndousse, et~al. 2022.
\newblock Red teaming language models to reduce harms: Methods, scaling behaviors, and lessons learned.
\newblock \emph{arXiv preprint arXiv:2209.07858}.

\bibitem[{Hendrycks et~al.(2021)Hendrycks, Burns, Basart, Zou, Mazeika, Song, and Steinhardt}]{mmlu}
Dan Hendrycks, Collin Burns, Steven Basart, Andy Zou, Mantas Mazeika, Dawn Song, and Jacob Steinhardt. 2021.
\newblock Measuring massive multitask language understanding.
\newblock In \emph{{International Conference on Learning Representations}}.

\bibitem[{Hu et~al.(2022)Hu, Shen, Wallis, Allen{-}Zhu, Li, Wang, Wang, and Chen}]{lora}
Edward~J. Hu, Yelong Shen, Phillip Wallis, Zeyuan Allen{-}Zhu, Yuanzhi Li, Shean Wang, Lu~Wang, and Weizhu Chen. 2022.
\newblock Lora: Low-rank adaptation of large language models.
\newblock In \emph{The Tenth International Conference on Learning Representations, {ICLR} 2022, Virtual Event, April 25-29, 2022}. OpenReview.net.

\bibitem[{Jin et~al.(2023)Jin, Yang, Chen, and Lu}]{genegpt}
Qiao Jin, Yifan Yang, Qingyu Chen, and Zhiyong Lu. 2023.
\newblock Genegpt: Augmenting large language models with domain tools for improved access to biomedical information.
\newblock \emph{ArXiv}.

\bibitem[{Kirkpatrick et~al.(2017)Kirkpatrick, Pascanu, Rabinowitz, Veness, Desjardins, Rusu, Milan, Quan, Ramalho, Grabska-Barwinska et~al.}]{ewc}
James Kirkpatrick, Razvan Pascanu, Neil Rabinowitz, Joel Veness, Guillaume Desjardins, Andrei~A Rusu, Kieran Milan, John Quan, Tiago Ramalho, Agnieszka Grabska-Barwinska, et~al. 2017.
\newblock Overcoming catastrophic forgetting in neural networks.
\newblock \emph{Proceedings of the national academy of sciences}, 114(13):3521--3526.

\bibitem[{Li et~al.(2023)Li, Zhang, Dubois, Taori, Gulrajani, Guestrin, Liang, and Hashimoto}]{alpaca_eval}
Xuechen Li, Tianyi Zhang, Yann Dubois, Rohan Taori, Ishaan Gulrajani, Carlos Guestrin, Percy Liang, and Tatsunori~B. Hashimoto. 2023.
\newblock Alpacaeval: An automatic evaluator of instruction-following models.
\newblock \url{https://github.com/tatsu-lab/alpaca_eval}.

\bibitem[{Lin et~al.(2021)Lin, Hilton, and Evans}]{truthfulqa}
Stephanie Lin, Jacob Hilton, and Owain Evans. 2021.
\newblock \href {http://arxiv.org/abs/2109.07958} {Truthfulqa: Measuring how models mimic human falsehoods}.

\bibitem[{Lopez-Paz and Ranzato(2017)}]{gem}
David Lopez-Paz and Marc'Aurelio Ranzato. 2017.
\newblock Gradient episodic memory for continual learning.
\newblock In \emph{{Advances in Neural Information Processing Systems}}, volume~30.

\bibitem[{Luo et~al.(2023{\natexlab{a}})Luo, Sun, Xu, Zhao, Lou, Tao, Geng, Lin, Chen, and Zhang}]{wizardmath}
Haipeng Luo, Qingfeng Sun, Can Xu, Pu~Zhao, Jianguang Lou, Chongyang Tao, Xiubo Geng, Qingwei Lin, Shifeng Chen, and Dongmei Zhang. 2023{\natexlab{a}}.
\newblock Wizardmath: Empowering mathematical reasoning for large language models via reinforced evol-instruct.
\newblock \emph{arXiv preprint arXiv:2308.09583}.

\bibitem[{Luo et~al.(2023{\natexlab{b}})Luo, Yang, Meng, Li, Zhou, and Zhang}]{luo2023empirical}
Yun Luo, Zhen Yang, Fandong Meng, Yafu Li, Jie Zhou, and Yue Zhang. 2023{\natexlab{b}}.
\newblock An empirical study of catastrophic forgetting in large language models during continual fine-tuning.
\newblock \emph{arXiv preprint arXiv:2308.08747}.

\bibitem[{Luo et~al.(2024)Luo, Xu, Zhao, Sun, Geng, Hu, Tao, Ma, Lin, and Jiang}]{wizardcoder}
Ziyang Luo, Can Xu, Pu~Zhao, Qingfeng Sun, Xiubo Geng, Wenxiang Hu, Chongyang Tao, Jing Ma, Qingwei Lin, and Daxin Jiang. 2024.
\newblock Wizardcoder: Empowering code large language models with evol-instruct.
\newblock In \emph{{International Conference on Learning Representations}}.

\bibitem[{Madaan et~al.(2024)Madaan, Tandon, Gupta, Hallinan, Gao, Wiegreffe, Alon, Dziri, Prabhumoye, Yang et~al.}]{selfrefine}
Aman Madaan, Niket Tandon, Prakhar Gupta, Skyler Hallinan, Luyu Gao, Sarah Wiegreffe, Uri Alon, Nouha Dziri, Shrimai Prabhumoye, Yiming Yang, et~al. 2024.
\newblock Self-refine: Iterative refinement with self-feedback.
\newblock In \emph{{Advances in Neural Information Processing Systems}}, volume~36.

\bibitem[{Mallya and Lazebnik(2018)}]{packnet}
Arun Mallya and Svetlana Lazebnik. 2018.
\newblock Packnet: Adding multiple tasks to a single network by iterative pruning.
\newblock In \emph{{Proceedings of the IEEE/CVF Conference on Computer Vision and Pattern Recognition}}, pages 7765--7773.

\bibitem[{{Meta AI}(2024)}]{llama3}
{Meta AI}. 2024.
\newblock Introducing meta llama 3: The most capable openly available llm to date.
\newblock \url{https://ai.meta.com/blog/meta-llama-3}.

\bibitem[{Ouyang et~al.(2022{\natexlab{a}})Ouyang, Wu, Jiang, Almeida, Wainwright, Mishkin, Zhang, Agarwal, Slama, Ray et~al.}]{DBLP:conf/nips/Ouyang0JAWMZASR22}
Long Ouyang, Jeffrey Wu, Xu~Jiang, Diogo Almeida, Carroll Wainwright, Pamela Mishkin, Chong Zhang, Sandhini Agarwal, Katarina Slama, Alex Ray, et~al. 2022{\natexlab{a}}.
\newblock Training language models to follow instructions with human feedback.
\newblock In \emph{{Advances in Neural Information Processing Systems}}.

\bibitem[{Ouyang et~al.(2022{\natexlab{b}})Ouyang, Wu, Jiang, Almeida, Wainwright, Mishkin, Zhang, Agarwal, Slama, Ray, Schulman, Hilton, Kelton, Miller, Simens, Askell, Welinder, Christiano, Leike, and Lowe}]{instructgpt}
Long Ouyang, Jeffrey Wu, Xu~Jiang, Diogo Almeida, Carroll~L. Wainwright, Pamela Mishkin, Chong Zhang, Sandhini Agarwal, Katarina Slama, Alex Ray, John Schulman, Jacob Hilton, Fraser Kelton, Luke Miller, Maddie Simens, Amanda Askell, Peter Welinder, Paul~F. Christiano, Jan Leike, and Ryan Lowe. 2022{\natexlab{b}}.
\newblock \href {http://papers.nips.cc/paper\_files/paper/2022/hash/b1efde53be364a73914f58805a001731-Abstract-Conference.html} {Training language models to follow instructions with human feedback}.
\newblock In \emph{Advances in Neural Information Processing Systems 35: Annual Conference on Neural Information Processing Systems 2022, NeurIPS 2022, New Orleans, LA, USA, November 28 - December 9, 2022}.

\bibitem[{Patil et~al.(2023)Patil, Zhang, Wang, and Gonzalez}]{openfunction}
Shishir~G. Patil, Tianjun Zhang, Xin Wang, and Joseph~E. Gonzalez. 2023.
\newblock Gorilla: Large language model connected with massive apis.
\newblock In \emph{arXiv preprint arXiv:2305.15334}.

\bibitem[{Pelrine et~al.(2023)Pelrine, Taufeeque, Zajac, McLean, and Gleave}]{pelrine2023exploiting}
Kellin Pelrine, Mohammad Taufeeque, Michal Zajac, Euan McLean, and Adam Gleave. 2023.
\newblock Exploiting novel gpt-4 apis.
\newblock \emph{arXiv preprint arXiv:2312.14302}.

\bibitem[{Perez et~al.(2022)Perez, Huang, Song, Cai, Ring, Aslanides, Glaese, McAleese, and Irving}]{redteam1}
Ethan Perez, Saffron Huang, Francis Song, Trevor Cai, Roman Ring, John Aslanides, Amelia Glaese, Nat McAleese, and Geoffrey Irving. 2022.
\newblock Red teaming language models with language models.
\newblock \emph{arXiv preprint arXiv:2202.03286}.

\bibitem[{Qi et~al.(2024)Qi, Zeng, Xie, Chen, Jia, Mittal, and Henderson}]{qi2023fine}
Xiangyu Qi, Yi~Zeng, Tinghao Xie, Pin-Yu Chen, Ruoxi Jia, Prateek Mittal, and Peter Henderson. 2024.
\newblock Fine-tuning aligned language models compromises safety, even when users do not intend to!
\newblock In \emph{{International Conference on Learning Representations}}.

\bibitem[{Reimers and Gurevych(2019)}]{sentencebert}
Nils Reimers and Iryna Gurevych. 2019.
\newblock \href {https://doi.org/10.18653/v1/D19-1410} {Sentence-{BERT}: Sentence embeddings using {S}iamese {BERT}-networks}.
\newblock In \emph{Proceedings of the 2019 Conference on Empirical Methods in Natural Language Processing and the 9th International Joint Conference on Natural Language Processing (EMNLP-IJCNLP)}, pages 3982--3992, Hong Kong, China. Association for Computational Linguistics.

\bibitem[{Roziere et~al.(2023)Roziere, Gehring, Gloeckle, Sootla, Gat, Tan, Adi, Liu, Remez, Rapin et~al.}]{codellama}
Baptiste Roziere, Jonas Gehring, Fabian Gloeckle, Sten Sootla, Itai Gat, Xiaoqing~Ellen Tan, Yossi Adi, Jingyu Liu, Tal Remez, J{\'e}r{\'e}my Rapin, et~al. 2023.
\newblock Code llama: Open foundation models for code.
\newblock \emph{arXiv preprint arXiv:2308.12950}.

\bibitem[{Sakaguchi et~al.(2021)Sakaguchi, Bras, Bhagavatula, and Choi}]{winogrande}
Keisuke Sakaguchi, Ronan~Le Bras, Chandra Bhagavatula, and Yejin Choi. 2021.
\newblock Winogrande: An adversarial winograd schema challenge at scale.
\newblock \emph{Communications of the ACM}, 64(9):99--106.

\bibitem[{Scialom et~al.(2022)Scialom, Chakrabarty, and Muresan}]{scialom2022fine}
Thomas Scialom, Tuhin Chakrabarty, and Smaranda Muresan. 2022.
\newblock Fine-tuned language models are continual learners.
\newblock In \emph{Proceedings of the 2022 Conference on Empirical Methods in Natural Language Processing}, pages 6107--6122.

\bibitem[{Sun et~al.(2023)Sun, Shen, Zhou, Zhang, Chen, Cox, Yang, and Gan}]{selfalign}
Zhiqing Sun, Yikang Shen, Qinhong Zhou, Hongxin Zhang, Zhenfang Chen, David Cox, Yiming Yang, and Chuang Gan. 2023.
\newblock Principle-driven self-alignment of language models from scratch with minimal human supervision.
\newblock In \emph{{Advances in Neural Information Processing Systems}}.

\bibitem[{Taori et~al.(2023)Taori, Gulrajani, Zhang, Dubois, Li, Guestrin, Liang, and Hashimoto}]{alpaca}
Rohan Taori, Ishaan Gulrajani, Tianyi Zhang, Yann Dubois, Xuechen Li, Carlos Guestrin, Percy Liang, and Tatsunori~B. Hashimoto. 2023.
\newblock Stanford alpaca: An instruction-following llama model.
\newblock \url{https://github.com/tatsu-lab/stanford_alpaca}.

\bibitem[{Teknium(2023)}]{openhermes}
Teknium. 2023.
\newblock \href {https://huggingface.co/datasets/teknium/OpenHermes} {Openhermes dataset}.

\bibitem[{Touvron et~al.(2023)Touvron, Martin, Stone, Albert, Almahairi, Babaei, Bashlykov, Batra, Bhargava, Bhosale et~al.}]{llama2}
Hugo Touvron, Louis Martin, Kevin Stone, Peter Albert, Amjad Almahairi, Yasmine Babaei, Nikolay Bashlykov, Soumya Batra, Prajjwal Bhargava, Shruti Bhosale, et~al. 2023.
\newblock Llama 2: Open foundation and fine-tuned chat models.
\newblock \emph{arXiv preprint arXiv:2307.09288}, abs/2307.09288.

\bibitem[{Wang et~al.(2023)Wang, Ivison, Dasigi, Hessel, Khot, Chandu, Wadden, MacMillan, Smith, Beltagy, and Hajishirzi}]{tulu}
Yizhong Wang, Hamish Ivison, Pradeep Dasigi, Jack Hessel, Tushar Khot, Khyathi~Raghavi Chandu, David Wadden, Kelsey MacMillan, Noah~A. Smith, Iz~Beltagy, and Hannaneh Hajishirzi. 2023.
\newblock \href {https://doi.org/10.48550/ARXIV.2306.04751} {How far can camels go? exploring the state of instruction tuning on open resources}.
\newblock \emph{CoRR}, abs/2306.04751.

\bibitem[{Wang et~al.(2022)Wang, Kordi, Mishra, Liu, Smith, Khashabi, and Hajishirzi}]{selfinstruct}
Yizhong Wang, Yeganeh Kordi, Swaroop Mishra, Alisa Liu, Noah~A Smith, Daniel Khashabi, and Hannaneh Hajishirzi. 2022.
\newblock Self-instruct: Aligning language model with self generated instructions.
\newblock \emph{arXiv preprint arXiv:2212.10560}.

\bibitem[{Wei et~al.(2023)Wei, Wang, Liu, Ding, and Zhang}]{magicoder}
Yuxiang Wei, Zhe Wang, Jiawei Liu, Yifeng Ding, and Lingming Zhang. 2023.
\newblock Magicoder: Source code is all you need.
\newblock \emph{arXiv preprint arXiv:2312.02120}.

\bibitem[{Wu et~al.(2023)Wu, Irsoy, Lu, Dabravolski, Dredze, Gehrmann, Kambadur, Rosenberg, and Mann}]{financellm}
Shijie Wu, Ozan Irsoy, Steven Lu, Vadim Dabravolski, Mark Dredze, Sebastian Gehrmann, Prabhanjan Kambadur, David Rosenberg, and Gideon Mann. 2023.
\newblock Bloomberggpt: A large language model for finance.
\newblock \emph{arXiv preprint arXiv:2303.17564}.

\bibitem[{Xu et~al.(2024)Xu, Sun, Zheng, Geng, Zhao, Feng, Tao, and Jiang}]{wizardlm}
Can Xu, Qingfeng Sun, Kai Zheng, Xiubo Geng, Pu~Zhao, Jiazhan Feng, Chongyang Tao, and Daxin Jiang. 2024.
\newblock Wizardlm: Empowering large language models to follow complex instructions.
\newblock In \emph{{International Conference on Learning Representations}}.

\bibitem[{Yang et~al.(2023)Yang, Wang, Zhang, Petzold, Wang, Zhao, and Lin}]{shallowalign}
Xianjun Yang, Xiao Wang, Qi~Zhang, Linda Petzold, William~Yang Wang, Xun Zhao, and Dahua Lin. 2023.
\newblock Shadow alignment: The ease of subverting safely-aligned language models.
\newblock \emph{arXiv preprint arXiv:2310.02949}.

\bibitem[{Yuan et~al.(2024)Yuan, Pang, Cho, Sukhbaatar, Xu, and Weston}]{selfreward}
Weizhe Yuan, Richard~Yuanzhe Pang, Kyunghyun Cho, Sainbayar Sukhbaatar, Jing Xu, and Jason Weston. 2024.
\newblock Self-rewarding language models.
\newblock \emph{arXiv preprint arXiv:2401.10020}.

\bibitem[{Zellers et~al.(2019)Zellers, Holtzman, Bisk, Farhadi, and Choi}]{hellaswag}
Rowan Zellers, Ari Holtzman, Yonatan Bisk, Ali Farhadi, and Yejin Choi. 2019.
\newblock Hellaswag: Can a machine really finish your sentence?
\newblock In \emph{Proceedings of the 57th Annual Meeting of the Association for Computational Linguistics}.

\bibitem[{Zhan et~al.(2023)Zhan, Fang, Bindu, Gupta, Hashimoto, and Kang}]{zhan2023removing}
Qiusi Zhan, Richard Fang, Rohan Bindu, Akul Gupta, Tatsunori Hashimoto, and Daniel Kang. 2023.
\newblock Removing rlhf protections in gpt-4 via fine-tuning.
\newblock \emph{arXiv preprint arXiv:2311.05553}.

\bibitem[{Zhang et~al.(2023)Zhang, Zhang, Li, and Smola}]{autocot}
Zhuosheng Zhang, Aston Zhang, Mu~Li, and Alex Smola. 2023.
\newblock Automatic chain of thought prompting in large language models.
\newblock In \emph{{International Conference on Learning Representations}}.

\bibitem[{Zhou et~al.(2023)Zhou, Liu, Xu, Iyer, Sun, Mao, Ma, Efrat, Yu, Yu et~al.}]{lima}
Chunting Zhou, Pengfei Liu, Puxin Xu, Srini Iyer, Jiao Sun, Yuning Mao, Xuezhe Ma, Avia Efrat, Ping Yu, Lili Yu, et~al. 2023.
\newblock Lima: Less is more for alignment.
\newblock \emph{arXiv preprint arXiv:2305.11206}.

\bibitem[{Zou et~al.(2023)Zou, Wang, Kolter, and Fredrikson}]{advbench}
Andy Zou, Zifan Wang, J~Zico Kolter, and Matt Fredrikson. 2023.
\newblock Universal and transferable adversarial attacks on aligned language models.
\newblock \emph{arXiv preprint arXiv:2307.15043}.

\end{thebibliography}

\clearpage
\appendix
\section{Experimental Details}
\label{app:details}
Throughout most experiments, we applied fine-tuning to Llama-2-7b-chat with the Low-Rank Adaptation (LoRA) technique~\citep{lora}. The query and value matrices of the LoRA were tuned with a rank of $r=8$. We adhered to the default configuration settings of Llama2. The learning rate was initiated at $1 \times 10^{-4}$ and progressively decayed to zero following a cosine annealing schedule. and the batch size was set to $8$. 

We randomly sampled a subset of 2,000 examples and conducted fine-tuning for 2 epochs for the Alpaca~\cite{alpaca}, Dolly~\cite{dolly}, and MagiCoder~\cite{magicoder} datasets. We sampled 20,000 examples for the OpenHermes dataset and train 2 epochs. For GSM8K~\cite{gsm8k}, LIMA~\cite{lima} and OpenFunctions~\cite{openfunction} datasets, we fine-tune on the entire train set. We train LIMA for 2 epochs and the other two datasets for 5 epochs.

To assess the general helpfulness of models, we employ the AlpacaEval framework~\footnote{\url{https://github.com/tatsu-lab/alpaca\_eval}}, with GPT-4 serving as the evaluator. The evaluations of benchmarks in OpenLLM leaderboard are implemented via the lm-evaluation-harness~\footnote{\url{https://github.com/EleutherAI/lm-evaluation-harness}}. Furthermore, we evaluate the coding capabilities using the HumanEval dataset~\cite{humaneval}, leveraging the bigcode-evaluation-harness project\footnote{\url{https://github.com/bigcode-project/bigcode-evaluation-harness}}.

\section{Templates and Examples}
\label{app:templates}
This section provides templates used in our experiments and some illustrative examples of distillation on each dataset.

In most of our experiments, we use the standard alpaca~\cite{alpaca} template for both fine-tuning and prediction, as presented in Figure~\ref{fig:template_alpaca}.

To enhance the reasoning abilities, we slightly modify the standard Alpaca template for datasets regarding reasoning, namely GSM8K and MultiArith. The template used for training and distillation are presented in Figure~\ref{fig:template_gsm8k_train} and Figure~\ref{fig:template_gsm8k_distill} respectively.

\begin{figure}[ht]
    \centering
    \RoundBox{
    Below are an instruction that describes a task. Write a response that appropriately completes the request.
    
    \vspace{\baselineskip}  
    \#\#\# Instruction:
    
    \textit{\{instruction\}}
    
    \vspace{\baselineskip}
    \#\#\# Response:
    }    
    \caption{The standard alpaca template. This template is used for both training and evaluation in most experiments.}
    \label{fig:template_alpaca}
\end{figure}

\begin{figure}
    \centering
    \RoundBox{
    You are an expert in math. Below is a math question. Write a response that appropriately answers the question.                                       
    \vspace{\baselineskip}  
    
    \textit{\{instruction\}}
    }
    \caption{The template used for training on the GSM8K dataset.}
    \label{fig:template_gsm8k_train}
\end{figure}

\begin{figure}
    \centering
    \centering
    \RoundBox{
    You are an expert in math. Below are a math question and its reference answer. Refer to the reference answer and write a response that appropriately answers the question.
    
    \vspace{\baselineskip}  
    
    \textit{\{instruction\}}

    \vspace{\baselineskip}
    
    \textit{\{response\}}

    Great! Let's think step by step.
    }
    \caption{The template used for distilling on the GSM8K dataset.}
    \label{fig:template_gsm8k_distill}
\end{figure}

To make the extraction of final answer easier for mathematical reasoning datasets, we specify the format of final answer during evaluation. The template used for evaluation is presented in Figure~\ref{fig:template_gsm8k_evaluation}.

\begin{figure}
    \centering
    \RoundBox{
    You are an expert in math. Below is a math question. Write a response that appropriately answers the question. Your final answer should be an integer at the end of your response, formatted as: The answer is \{answer\}.
    
    \vspace{\baselineskip}  
    
    \textit{\{instruction\}}
    }
    \caption{The template used for evaluation on the GSM8K and MultiArith datasets.}
    \label{fig:template_gsm8k_evaluation}
\end{figure}

Figures~\ref{fig:example-alpaca}, \ref{fig:example-dolly}, \ref{fig:example-gsm8k}, \ref{fig:example-openfunction}, \ref{fig:example-lima} present examples of distilled data point on each dataset.

\begin{figure*}[ht]
    \centering
    \BigExampleBox{
    \textbf{Instruction:}
    
    Name three types of machine learning algorithms.
    
    \vspace{\baselineskip}
    \textbf{Original Answer:}
    
    Three types of machine learning algorithms are supervised learning, unsupervised learning, and reinforcement learning.
    
    \vspace{\baselineskip}
    \textbf{Distilled Response:}

    I can name three types of machine learning algorithms as follows:

    1. Supervised Learning: This type of algorithm involves training a machine learning model on labeled data, where the target output is already known. The model learns to predict the target output based on the input data, and can be used for tasks such as image classification, speech recognition, and sentiment analysis.
    
    2. Unsupervised Learning: This type of algorithm involves training a machine learning model on unlabeled data, where there is no target output. The model learns patterns and relationships in the data without any prior knowledge of the output. Examples of unsupervised learning tasks include clustering, dimensionality reduction, and anomaly detection.
    
    3. Reinforcement Learning: This type of algorithm involves training a machine learning model to make a series of decisions in an environment in order to maximize a reward. The model learns through trial and error, and the reward serves as feedback to guide the learning process. Examples of reinforcement learning tasks include game playing, robotics, and autonomous driving.    
    }    
    \caption{Example of distilled data point on the Alpaca dataset.}
    \label{fig:example-alpaca}
\end{figure*}


\begin{figure*}
    \centering
    \BigExampleBox{
    \textbf{Instruction:}
    
    Classify the following as a 'Country', 'State' or 'City': California, Sacramento, Chicago, Singapore, New Delhi, London, Tennessee, New York
    
    \vspace{\baselineskip}
    \textbf{Original Answer:}
    
    California: State
    
    Sacramento: City 
    
    Chicago: City
    
    Singapore: Both Country and City
    
    New Delhi: City
    
    London: City
    
    Tennessee: City
    
    New York: Both State and City
    
    \vspace{\baselineskip}
    \textbf{Distilled Response:}
    
    I classify the following as:
    
    * California: State
    
    * Sacramento: City
    
    * Chicago: City
    
    * Singapore: Both Country and City
    
    * New Delhi: City
    
    * London: City
    
    * Tennessee: City
    
    * New York: Both State and City
    }
    \caption{Example of distilled data point on the Dolly dataset.}
    \label{fig:example-dolly}
\end{figure*}

\begin{figure*}
    \centering
    \BigExampleBox{
    \textbf{Instruction:}
    
    Lisa, Jack, and Tommy earned \$60 from washing cars all week. However, half of the \$60 was earned by Lisa. Tommy earned half of what Lisa earned. 
    
    How much more money did Lisa earn than Tommy?
    
    \vspace{\baselineskip}
    \textbf{Original Answer:}
    
    Lisa earned \$60 * 1/2 = \$30.
    
    Tommy earned \$30 * 1/2 = \$15.
    
    Lisa earned \$30 - \$15 = \$15 more than Tommy.
    
    The answer is 15.
    
    \vspace{\baselineskip}
    \textbf{Distilled Response:}
    
    Here is the response:
    
    Lisa earned \$60 * 1/2 = \$30, and Tommy earned \$30 * 1/2 = \$15. So, Lisa earned \$30 - \$15 = \$15 more than Tommy. Therefore, the answer is \$15.
    }
    \caption{Example of distilled data point on the GSM8K dataset.}
    \label{fig:example-gsm8k}
\end{figure*}

\begin{figure*}[t]
    \centering
        \BigExampleBox{
    \textbf{Instruction:}
    
    Construct a function call to answer the given question using keyword arguments based on the provided specifications:
    
    \vspace{\baselineskip}
    Question: I want to describe the Studio Lifecycle Configuration named 'my-lc-config' on AWS SageMaker.

    \vspace{\baselineskip}
    Function to be used: aws.sagemaker.describe\_studio\_lifecycle\_config

    \vspace{\baselineskip}
    Description: Describes the Studio Lifecycle Configuration.

    \vspace{\baselineskip}
    Parameters:
    
    1. studio\_lifecycle\_config\_name - The name of the Studio Lifecycle Configuration to describe.
    
    Format your response as a single string using keyword arguments.
    
    \vspace{\baselineskip}
    \textbf{Original Answer:}
    
    aws.sagemaker.describe\_studio\_lifecycle\_config(studio\_lifecycle\_config\_name=``my-lc-config'')
    
    \vspace{\baselineskip}
    \textbf{Distilled Response:}
    
    aws.sagemaker.describe\_studio\_lifecycle\_config(studio\_lifecycle\_config\_name="my-lc-config")

    \vspace{\baselineskip}
    \#\#\# Explanation:

    The function call uses the `studio\_lifecycle\_config\_name` parameter to specify the name of the Studio Lifecycle Configuration to describe, which in this case is "my-lc-config". The `aws.sagemaker.describe\_studio\_lifecycle\_config` function is then called with this parameter to retrieve the description of the specified Studio Lifecycle Configuration.
    }
    \caption{Example of distilled data point on the OpenFuctions dataset.}
    \label{fig:example-openfunction}
\end{figure*}

\begin{figure*}
    \centering
    \BigExampleBox{
    \textbf{Instruction:}
    
    How slicing in Python works? Please include references where appropriate.
    
    \vspace{\baselineskip}
    \textbf{Original Answer:}
    
    The syntax is:
    
    ```
    
    a[start:stop]  \# items start through stop-1
    
    a[start:]      \# items start through the rest of the array
    
    a[:stop]       \# items from the beginning through stop-1
    
    a[:]           \# a copy of the whole array
    
    ```

    There is also the `step` value, which can be used with any of the above:
    ```
    a[start:stop:step] \# start through not past stop, by step
    ```
    The key point to remember is that the `:stop` value represents the first value that is not in the selected slice. So, the difference between `stop` and `start` is the number of elements selected (if `step` is 1, the default)
    
    $\dots$

    \vspace{\baselineskip}
    While the ```:```-based notation is very helpful for simple slicing, the explicit use of ```slice()``` objects simplifies the programmatic generation of slicing.
    
    \vspace{\baselineskip}
    \textbf{Distilled Response:}
    
    Slicing in Python is a powerful feature that allows you to extract a subset of elements from a sequence (such as a list, tuple, or string) based on a specified range and step. The syntax for slicing is:
    
    ```
    
    a[start:stop]  \# items start through stop-1
    
    a[start:]      \# items start through the rest of the array
    
    a[:stop]       \# items from the beginning through stop-1
    
    a[:]           \# a copy of the whole array
    
    ```

    In addition to the basic syntax, there are several other features of slicing in Python:

    * **Negative indices**: If `start` or `stop` is a negative number, it counts from the end of the array instead of the beginning. For example, `a[-2]` returns the last two elements of the array, while `a[-3]` returns the last three elements.
    
    \vspace{\baselineskip}
    * **Step**: The `step` value can be used to specify the distance between elements in the slice. For example, `a[1:3:2]` returns the elements at indices 1, 3, and 5 of the array.
    
    \vspace{\baselineskip}
    * **Optional arguments**: If `stop` is not specified, it defaults to the end of the array. Similarly, if `step` is not specified, it defaults to 1
    
    $\dots$

    \vspace{\baselineskip}
    In summary, slicing in Python is a powerful feature that allows you to extract a subset of elements from a sequence based on a specified range and step. The basic syntax is simple and intuitive, but there are several other features and options that can be used to customize the slicing operation
    
    }
    \caption{Example of distilled data point on the LIMA dataset.}
    \label{fig:example-lima}
\end{figure*}

\end{document}